\documentclass{article} 
\PassOptionsToPackage{table}{xcolor}
\usepackage{iclr2026_conference,times}

\usepackage{amsmath,amsfonts,bm}

\def\eqref#1{equation~\ref{#1}}

\def\1{\bm{1}}

\DeclareMathAlphabet{\mathsfit}{\encodingdefault}{\sfdefault}{m}{sl}
\SetMathAlphabet{\mathsfit}{bold}{\encodingdefault}{\sfdefault}{bx}{n}

\usepackage{hyperref}
\usepackage{url}
\usepackage{booktabs}
\usepackage{graphicx}
\usepackage{multirow}
\usepackage{amsmath,amssymb,amsthm}
\usepackage{xcolor}
\definecolor{rowhl}{RGB}{232,242,255}
\usepackage{float}
\usepackage{algorithm}
\usepackage{algorithmic}
\usepackage{fancyvrb}
\fvset{fontsize=\small}
\usepackage[most]{tcolorbox}
\newtcolorbox{promptbox}[1][]{
  enhanced, colback=gray!5, colframe=gray!60,
  fonttitle=\bfseries\small, title=#1,
  boxrule=0.5pt, arc=2pt, left=6pt, right=6pt, top=4pt, bottom=4pt}
\makeatletter
\renewcommand\paragraph{\@startsection{paragraph}{4}{\z@}{0.5ex plus 0.2ex minus 0.1ex}{-1em}{\normalfont\normalsize\bfseries}}
\makeatother

\title{LLMs as a Jury: Cross-Model Consensus Can Outperform Process Reward Models for LLM Reasoning}

\author{Ning Liu \\
Independent Researcher \\
\texttt{ningliu@umich.edu} }

\iclrpreprintcopy
\begin{document}
\maketitle

\begin{abstract}
Selecting the correct answer from a pool of candidate reasoning chains is the engine of test-time scaling, yet the standard selectors each carry a cost: self-consistency inherits the errors of the single model it resamples, and trained reward models need labeled data and transfer poorly off-distribution. We study a third signal, free at inference time: \textbf{cross-model consensus}, the degree to which independently trained models, each solving the problem once, agree on a final answer. We treat the panel as an \textbf{LLM-jury}, in which the verification signal is the structure of agreement itself, with no model scoring another's work. Across seven benchmarks it selects correct answers better than self-consistency and far better than a model scoring its own candidates: on competition math it closes the entire gap to an oracle selector, while self-scoring closes almost none. The mechanism is error decorrelation: independently trained models err differently, so their wrong answers scatter while the correct one accumulates agreement. We make this precise with a parameter-free law, derived in closed form, that predicts consensus accuracy from three measured panel statistics to a mean absolute error of $0.03$ and exposes the method's ceiling: a \emph{shared-error floor} where models share a misconception, near zero on math but non-trivial on science. Against four trained verifiers spanning discriminative, outcome, and generative reward models, the free LLM-jury matches the strongest inside their math training domain and is the top selector outside it. Cross-model consensus is thus a verifier we can characterize in advance: a law that says when to trust it, and a floor that marks where it cannot.
\end{abstract}

\section{Introduction}
\label{sec:intro}

\begin{figure}[t]
\centering
\includegraphics[width=\linewidth]{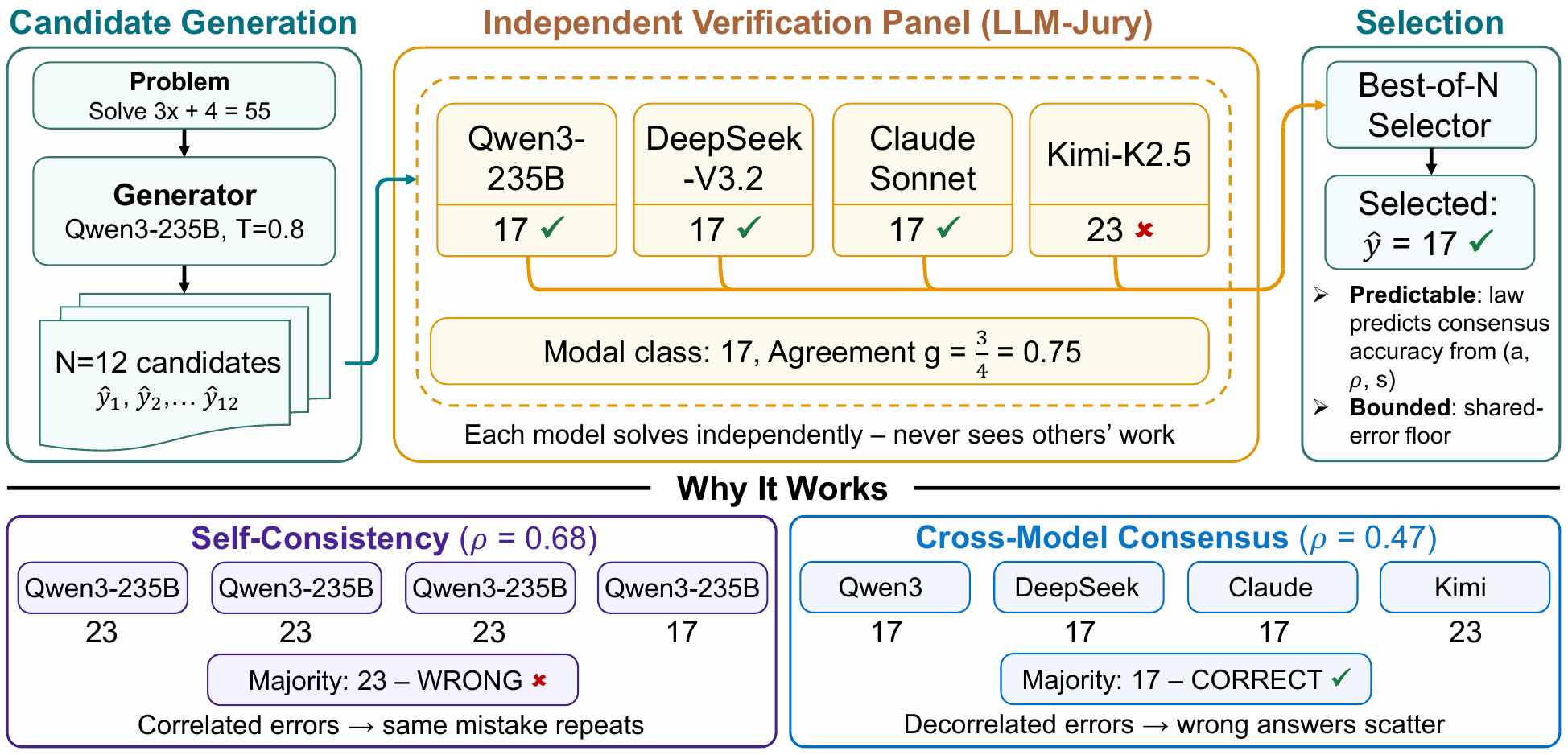}
\caption{\textbf{The LLM-jury.} A generator produces $N$ candidate solutions. An \emph{independent verification panel} of cross-family models each solves the same problem once, never seeing the candidates or one another's work, and the modal answer class is selected, with the agreement fraction $g$ as a free confidence signal. \emph{Why it works} (bottom): resampling one model (self-consistency) repeats correlated errors and can raise a wrong answer to the majority, whereas a decorrelated cross-model panel scatters its wrong answers so the correct one prevails. The signal is \emph{predictable} (a parameter-free law forecasts consensus accuracy from $(a,\rho,s)$) and \emph{bounded} (a shared-error floor).}
\label{fig:method}
\end{figure}

Modern reasoning systems spend compute at inference time by sampling many candidate solutions and selecting among them~\citep{wei2022chain,brown2024monkeys,snell2025scaling,muennighoff2025s1}. As the sample count grows a correct answer almost always appears somewhere in the pool, so the bottleneck shifts from generation to selection: the \emph{verifier} that picks which candidate to return increasingly bounds end-to-end accuracy~\citep{zhao2025sample,liu2025compute}. Yet the two verifiers that dominate practice each carry a structural cost. \emph{Self-consistency} takes a majority vote over repeated samples of a single model~\citep{wang2023selfconsistency}. Because every sample comes from the same model, it inherits that model's systematic errors, so a confidently wrong model errs on most samples and votes its own error into the majority. \emph{Trained reward models} (outcome and process reward models, or PRMs)~\citep{lightman2023verify,wang2024mathshepherd,zhang2025prmlessons,liu2025acemath} score candidates with a learned network, but require annotated training data and, as we show, transfer poorly outside the distribution they are trained on. Both leave a gap: a verifier that is at once \emph{label-free}, \emph{general}, and able to catch the errors a single model is blind to.

We study a third signal that fills this gap, needing neither extra samples of one model nor any training: \textbf{cross-model consensus}, the degree to which several \emph{independently trained} models, each solving the problem once, agree on a final answer. We call this panel an \textbf{LLM-jury} (Figure~\ref{fig:method}): like jurors who deliberate independently before a verdict, each model solves the problem on its own and the \emph{structure of their agreement} is the signal, a unanimous panel marking a trustworthy answer and a split one a problem to escalate. Crucially, our jurors never see the candidates or one another's work, unlike a panel of LLM \emph{judges} that read and score generations~\citep{verga2024jury}. This is why, as we show, a model scoring its own candidates is the weakest of the selectors we compare.

Agreement tracks correctness for a simple reason, the classical bias--variance--diversity logic of ensembles~\citep{krogh1994ensemble,dietterich2000ensemble} made concrete for reasoning: when models' mistakes are \emph{decorrelated}, wrong answers scatter while the correct one accumulates agreement, whereas resampling a single model repeats the \emph{same} mistake and manufactures a confident but wrong majority. The magnitude of this effect is large: on competition math, a model scoring its own twelve candidates captures essentially $0\%$ of the achievable selection gain over self-consistency, while three models that never see one another's work capture $100\%$ of it: the signal that separates a correct answer from a wrong one lives \emph{between} independently trained models and is unavailable to any single one. Recent multi-model methods~\citep{wang2024moa,jiang2023llmblender,li2024moreagents} exploit a version of this, but as an aggregation heuristic for a better answer. Our claim is not that ``voting helps'' (known) but that the LLM-jury is a \emph{predictable, training-free verifier with a quantified ceiling} that rivals trained reward models and generalizes where they do not. We establish this by isolating the scoring signal on a fixed candidate pool (so a better selector cannot be confused with better candidates), giving a theory that predicts when the signal can be trusted, and comparing it directly against the trained verifiers it aims to replace. Our core contributions are:
\begin{itemize}
\itemsep0.15em
\item We establish agreement as a verifier: on a fixed Best-of-$N$ pool across seven benchmarks, the LLM-jury selects correct answers better than self-consistency and a single-model LLM-verifier (up to $+20$ points on AIME, $p{=}0.001$), while the self-scoring verifier captures $\approx 0\%$ of the oracle gap, so a model cannot adjudicate its own candidates. We show the advantage is robust to the generator and persists even with a frontier-model panel (\S\ref{sec:verifier}).
\item We derive a closed-form, parameter-free law that predicts consensus accuracy and its full selective-prediction curve from measured panel statistics to a mean absolute error of $0.03$, validate it against simulation and under leave-one-benchmark-out transfer, and use it to pin the method's limit: a \emph{shared-error floor} of $\approx 0$ on math but non-zero on science, set by how alike models' errors are (\S\ref{sec:theory}).
\item We compare the jury head-to-head against four trained verifiers spanning discriminative PRMs, an outcome RM, and a generative verifier (Qwen2.5-Math-PRM 7B/72B, AceMath-72B, ThinkPRM-14B) on an identical trace pool: the free jury ties the strongest inside their math training domain (MATH-500) and leads outside it (GPQA). We further extend the signal to executable code and a compute-saving agreement cascade (\S\ref{sec:prm}).
\item We isolate error decorrelation, distinct from multi-model combination, as the operative mechanism, through matched-size, matched-accuracy experiments and a direct measurement of pairwise error correlation: four decorrelated models beat self-consistency at $32$ samples of one (\S\ref{sec:mechanism}).
\end{itemize}

\section{Related Work}
\label{sec:related}

\paragraph{Test-time scaling and sampling-based selection.}
Spending inference compute by sampling many candidates and selecting among them is now a central paradigm: repeated sampling expands coverage of correct answers~\citep{brown2024monkeys}, test-time compute can rival parameter scaling~\citep{snell2025scaling,muennighoff2025s1,liu2025compute}, and scaling \emph{verification} governs search returns more than scaling generation does~\citep{zhao2025sample}. Since reinforcement learning with verifiable rewards largely surfaces reasoning already reachable by sampling~\citep{yue2025rlvr}, a strong label-free \emph{selector} is the principal lever. This line asks how returns scale. We ask \emph{which signal} selects best, and contribute a training-free selector with a governing law.

\paragraph{Trained reward and verifier models.}
The dominant learned verifiers are outcome and process reward models: step-level supervision beats outcome-only supervision~\citep{lightman2023verify}, automatic labels remove the annotation cost~\citep{wang2024mathshepherd}, and these extend to frontier mathematical reward models~\citep{zhang2025prmlessons,liu2025acemath} and generative and reasoning variants~\citep{zhang2025genrm,khalifa2025thinkprm}. A training-free alternative instead prompts a capable model to judge candidates~\citep{zheng2023judge}. Each carries a structural cost: learned reward models require annotated data and stay domain-specialized, and a model adjudicating its \emph{own} candidates is a weak verifier. We show the LLM-jury matches or exceeds trained verifiers (discriminative PRMs, an outcome RM, a generative verifier) both inside and outside their math training domain, at no training cost (\S\ref{sec:prm},~\S\ref{sec:verifier}).

\paragraph{Multi-model inference and ensembling.}
Combining models is an established route to stronger outputs, via ranking and fusion~\citep{jiang2023llmblender}, layered refinement~\citep{wang2024moa}, debate~\citep{du2024debate}, or voting over many agents~\citep{li2024moreagents}, though the gains over a well-prompted single model remain contested~\citep{wang2024rethinking}. Most related, a \emph{panel} of models can replace a single judge for evaluation~\citep{verga2024jury}, its members \emph{scoring} a generation. All treat cross-model combination as an \emph{aggregation heuristic} for a better answer. We instead treat agreement as a \emph{measurement}: members independently \emph{solve} each problem, and we predict the reliability of their agreement from panel statistics and quantify its failure mode. This grounds the signal in the classical bias--variance--diversity account of ensembles, in which decorrelated errors drive the gain~\citep{krogh1994ensemble,dietterich2000ensemble,kuncheva2003diversity,breiman2001random,lakshminarayanan2017deepensembles}. We show the same holds across models (\S\ref{sec:mechanism}).

\paragraph{Self-consistency and selective prediction.}
Self-consistency~\citep{wang2023selfconsistency} aggregates repeated samples of one model by majority vote, with extensions to free-form generation~\citep{chen2023universalsc}. Its agreement rate is a widely used confidence signal, part of a broader literature on eliciting calibrated confidence from language models~\citep{kadavath2022know,tian2023justask,zhang2025hiddenstates}. Because all samples derive from one model, the signal inherits that model's systematic errors, the same limitation that constrains self-correction without an external signal~\citep{huang2023selfcorrection}. We replace within-model resampling with across-model agreement, which matched-budget, matched-accuracy experiments identify as the operative difference (\S\ref{sec:mechanism}), and draw a precise distinction: self-consistency is better \emph{calibrated} while the cross-model signal is more \emph{accurate}.

\section{Cross-Model Consensus as a Verifier}
\label{sec:method}
We formalise cross-model consensus as a selection verifier (\S\ref{sec:setup}), then derive a parameter-free law for its behavior from a few panel statistics and make explicit the error-decorrelation mechanism the law predicts to be decisive (\S\ref{sec:law}).

\subsection{Setup: consensus as a Best-of-\texorpdfstring{$N$}{N} selector}
\label{sec:setup}
Let a panel of $M$ independently trained models each produce a final answer to a problem, and let $\hat{y}_i$ be the answer of model $i$. Answers are compared by task-appropriate equivalence (symbolic equality for math, choice match for multiple choice; \S\ref{app:setup}), inducing a partition of $\{\hat{y}_1,\dots,\hat{y}_M\}$ into agreement classes. The \emph{cross-model consensus} answer is the modal class's value, and the \emph{agreement} $g\in(0,1]$ is the fraction of models in that class. The verifier problem is to select one answer from a candidate pool, and evaluating every verifier on the \emph{same} pool isolates the scoring signal.

We compare four selectors. \textbf{Self-consistency} samples one model $N$ times and takes the majority answer~\citep{wang2023selfconsistency}. \textbf{Cross-model consensus} takes the panel-majority answer over $M$ independent models (matched so $N{=}M$). The \textbf{single-model LLM-verifier} has one model score each candidate and selects the highest~\citep{zhang2025genrm}. \textbf{Oracle} selects any correct candidate if one exists, upper-bounding selection from the pool. Two regimes are of interest: (i) a fixed pool of $N$ candidates from one strong generator, where the selectors differ only in how they score the same candidates (\S\ref{sec:verifier}); and (ii) selection used directly to answer, where the panel-agreement $g$ doubles as a calibrated confidence for abstention (\S\ref{sec:theory}, Appendix~\ref{app:selective}).

\subsection{A parameter-free law for consensus behavior}
\label{sec:law}
Cross-model consensus is usually treated as a heuristic. We show its entire selective-prediction behavior follows from three measurable panel statistics: mean member accuracy $a$, mean pairwise \emph{error correlation} $\rho$, and the \emph{shared-misconception rate} $s$, the probability that, when a model is wrong, it produces the single attractor wrong answer that other erring models also tend to produce.

\paragraph{Generative model.} Per problem, draw a latent difficulty $p\sim\mathrm{Beta}(ak,(1{-}a)k)$, a Beta distribution with mean $a$ and concentration $k{=}1/\rho-1$ chosen so the induced pairwise correlation of model errors equals $\rho$ (independence as $\rho\!\to\!0$). Each model is correct with probability $p$. When wrong, it lands on the shared attractor with probability $s$ and on an idiosyncratic wrong answer otherwise. This is the minimal model that couples the two forces governing agreement: a shared difficulty that correlates errors, and a shared attractor that makes some errors agree. From $(a,\rho,s,M)$ it induces the joint distribution of (agreement level, consensus correctness), and hence the \textbf{consensus accuracy} $\Pr[\text{modal class correct}]$, the \textbf{selective-prediction curve} (accuracy when accepting the most-agreed fraction of problems), and the \textbf{shared-error floor} $\Pr[\text{all } M \text{ agree on one wrong answer}]$, the irreducible error no agreement signal can detect.

\paragraph{Closed-form law.} These quantities are not merely simulated: each is an exact function of $(a,\rho,s,M)$. Marginalizing the latent difficulty, the number of correct members follows a Beta-Binomial, and the shared-error floor admits the closed form
\begin{equation}
\label{eq:floor}
\mathrm{Floor}(a,\rho,s,M)\;=\;s^{M}\,\underbrace{\prod_{j=0}^{M-1}\frac{(1-a)k+j}{k+j}}_{\mathbb{E}[(1-p)^{M}]},\qquad k=\tfrac1\rho-1,
\end{equation}
the probability that all $M$ members err ($\mathbb{E}[(1-p)^M]$) times the probability they all land on the shared attractor ($s^M$). The consensus accuracy and the full selective-prediction curve have matching closed forms (Appendix~\ref{app:closedform}), which agree with Monte-Carlo simulation to within $0.002$ across all seven benchmarks. The law is \emph{parameter-free}: its inputs are \emph{measured} panel statistics (estimation detailed in Appendix~\ref{app:law}), never fit to the prediction target. Equation~\ref{eq:floor} is written for a single shared attractor (the interpretable case in which one misconception dominates and $s$ is a scalar), but in general the wrong-answer structure is a measured mass profile $(s_1,\dots,s_K)$ that reduces to the scalar $s$ when $K{=}1$. The predictions we report (Table~\ref{tab:law}, Figure~\ref{fig:law}) use this measured profile, which departs from the scalar form only where wrong answers concentrate on several attractors, as in many-option multiple choice (Appendix~\ref{app:closedform}). \S\ref{sec:theory} validates the law held-out.

\paragraph{Why decorrelation is decisive.}
\label{sec:mech-theory}
The law makes the mechanism explicit and separates two effects that prior aggregation work conflates. Consensus \emph{accuracy} rises as $\rho$ falls: with decorrelated errors, the modal class is dominated by the (shared) correct answer because wrong answers scatter, whereas correlated errors let a wrong answer accumulate a spurious majority. Consensus \emph{calibration}, how well the agreement level $g$ predicts whether the selected answer is correct, depends instead chiefly on member accuracy. The two axes come apart in practice (\S\ref{sec:mechanism}): a within-model panel (self-consistency) can be \emph{better calibrated} yet \emph{less accurate} than a decorrelated cross-model panel at matched member accuracy, because resampling one model reproduces its errors coherently, a sharp confidence signal even as those repeated errors cap accuracy. The decisive variable for selection accuracy is therefore error decorrelation; multi-model combination helps only insofar as it decorrelates errors. The shared-error floor is what remains when decorrelation cannot help, a property of the models' shared blind spots measured per domain.

\section{Experimental Setup}
\label{sec:experiments}
\paragraph{Models and panels.} Cross-model panels are drawn from a cross-family pool (Qwen3-235B, DeepSeek-V3.2, Claude Sonnet~4.6, and Kimi-K2.5), with a same-family Qwen size ladder (Qwen3-32B, Qwen3-Next-80B, Qwen3-235B) as a decorrelation control. Unless noted, the Best-of-$N$ generator is Qwen3-235B and the selection panel is the three remaining cross-family models plus the generator. The generator-robustness study (\S\ref{sec:verifier}) repeats the comparison with a DeepSeek-V3.2 generator. Decoding settings and serving infrastructure are deferred to Appendix~\ref{app:setup}.

\paragraph{Benchmarks.} We evaluate on seven benchmarks spanning four reasoning types: competition math (AIME-2024/2025~\citep{aime2024}, MATH-500~\citep{hendrycks2021math}), olympiad math (OlympiadBench~\citep{he2024olympiadbench}), grade-school math (GSM8K~\citep{cobbe2021gsm8k}), graduate science (GPQA~\citep{rein2024gpqa}), and broad knowledge (MMLU-Pro~\citep{wang2024mmlupro}). The code-domain study (\S\ref{sec:prm}) adds HumanEval+~\citep{liu2023evalplus}. Each is run at its full public size (MATH-500 $500$, GPQA Diamond $198$, OlympiadBench $674$, GSM8K $1319$, AIME-2024/2025 $30$ each), except MMLU-Pro, for which a fixed random sample of $1000$ (deterministic, seed $0$) already resolves the comparisons at the reported precision. All selectors see identical problems, so comparisons are paired, and per-table counts are stated in each table.

\paragraph{Grading and statistics.} A single harness scores every selector: symbolic/numeric equivalence via \texttt{math\_verify} and a boxed-answer parser for math, and letter-choice match for multiple choice. Significance uses one-sided paired bootstrap $p$-values ($10$K resamples), reported in the text and captions and pooling AIME-2024/2025 where noted for power, with per-cell $95\%$ intervals in Appendix~\ref{app:setup}.

\section{Experiments}
\label{sec:results}
We evaluate four claims: cross-model consensus is a strong Best-of-$N$ verifier (\S\ref{sec:verifier}); its behavior and ceiling match the parameter-free law (\S\ref{sec:theory}); it outperforms trained verifiers and generalizes beyond them (\S\ref{sec:prm}); and error decorrelation is the operative mechanism (\S\ref{sec:mechanism}).

\subsection{Cross-model consensus is a strong Best-of-\texorpdfstring{$N$}{N} verifier}
\label{sec:verifier}
On a \emph{fixed} pool of $N{=}12$ candidates from one strong generator (Qwen3-235B), where every selector scores the same candidates so differences reflect the signal alone, the LLM-jury is the strongest non-oracle selector on every unsaturated benchmark (Table~\ref{tab:verifier}). It beats self-consistency significantly on every benchmark with headroom (up to $+20.0$ on AIME-2024, $p{=}0.001$) and beats the single-model LLM-verifier on all seven (significantly on all but the small-sample $n{=}30$ AIME-2025, $p{\le}0.001$). Its one near-tie with self-consistency is GPQA ($+2.6$, not significant at $n{=}198$), the science regime whose shared-error floor the law predicts to be highest (\S\ref{sec:theory}). Yet even there it beats the single-model verifier decisively ($+9.6$, $p{<}0.001$).

\begin{table}[t]
\caption{\textbf{Best-of-$N$ selection accuracy} ($N{=}12$ candidates from Qwen3-235B, with all selectors choosing from the same pool so that differences reflect the scoring signal alone). Cross-model consensus is the strongest non-oracle selector on every benchmark, capturing $100\%$ of the oracle gap on AIME-24. Per-benchmark margins and significance are reported in the text (\S\ref{sec:verifier}).}
\label{tab:verifier}
\centering\small
\setlength{\tabcolsep}{5pt}
\begin{tabular}{@{}lccccccc@{}}
\toprule
\textbf{Selector} & \textbf{AIME-24} & \textbf{AIME-25} & \textbf{MATH} & \textbf{Olympiad} & \textbf{GPQA} & \textbf{MMLU-Pro} & \textbf{GSM8K} \\
& \footnotesize$n{=}30$ & \footnotesize$n{=}30$ & \footnotesize$n{=}500$ & \footnotesize$n{=}674$ & \footnotesize$n{=}198$ & \footnotesize$n{=}1000$ & \footnotesize$n{=}1319$ \\
\midrule
Self-consistency      & $36.7$ & $33.3$ & $95.2$ & $65.4$ & $33.3$ & $43.8$ & $96.7$ \\
LLM-verifier          & $30.0$ & $33.3$ & $93.2$ & $62.2$ & $26.3$ & $43.9$ & $95.9$ \\
\rowcolor{rowhl}
Cross-model consensus & $\mathbf{56.7}$ & $\mathbf{40.0}$ & $\mathbf{96.6}$ & $\mathbf{72.6}$ & $\mathbf{35.9}$ & $\mathbf{46.2}$ & $\mathbf{97.2}$ \\
\midrule
Oracle (upper bound)  & $56.7$ & $43.3$ & $98.6$ & $79.1$ & $53.0$ & $48.9$ & $97.9$ \\
\bottomrule
\end{tabular}
\end{table}

Two findings are salient. \emph{The single-model verifier is the weakest selector}: a model scoring its own candidates captures essentially none of the oracle gap above self-consistency, and is negative on OlympiadBench (Appendix~\ref{app:verifier}): direct evidence that a model cannot adjudicate its own outputs, the same blind-spot that undermines self-consistency. \emph{The cross-model gain tracks headroom}: it is largest where the oracle gap is widest and vanishes once the pool saturates, bounded by the recoverable error. The result is not ``voting helps'' but that \emph{cross-model} agreement is a markedly better selection signal than within-model agreement or a learned single-model score.

\paragraph{The advantage does not depend on the generator.} A natural concern is that Table~\ref{tab:verifier} favors the LLM-jury because the candidate pool comes from one specific generator (Qwen3-235B). Repeating the entire seven-benchmark comparison with a different-family generator (DeepSeek-V3.2, with the remaining cross-family models as the panel) reproduces the result: the LLM-jury beats the single-model LLM-verifier on all seven benchmarks ($p{\le}0.016$) and beats self-consistency on every unsaturated one (AIME-2024 $+20.0$, $p{=}0.001$; OlympiadBench $+4.8$, $p{<}0.001$; GPQA $+6.6$, $p{=}0.006$; MMLU-Pro $+3.3$, $p{<}0.001$), tying only on saturated math and small-sample AIME-2025, the same no-headroom regimes as in Table~\ref{tab:verifier}. The selection-signal advantage is thus a property of the cross-model signal and does not depend on which generator supplies the pool (Appendix~\ref{app:gen2}).

\subsection{The parameter-free law predicts consensus behavior and its ceiling}
\label{sec:theory}
\begin{figure}[t]
\centering
\includegraphics[width=0.95\linewidth]{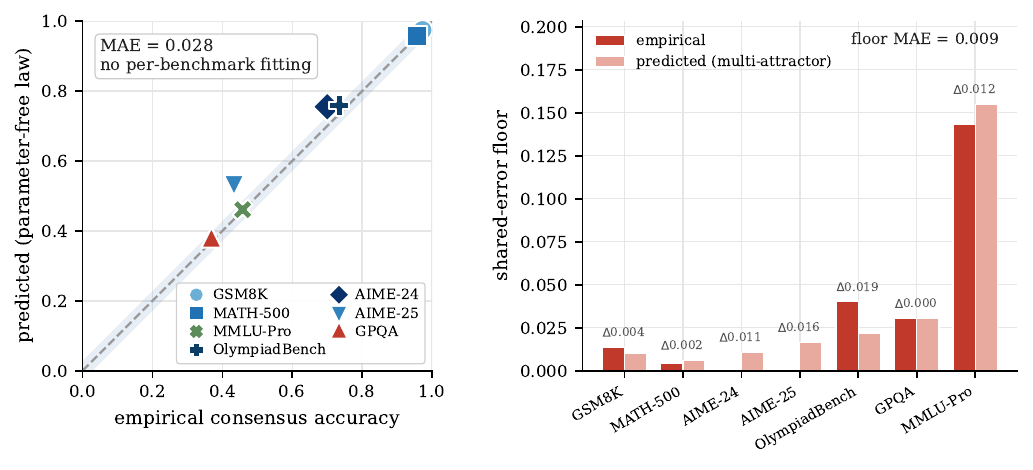}
\caption{\textbf{The parameter-free law.} Left: predicted vs.\ empirical consensus accuracy across seven benchmarks. Points lie near the identity line (mean abs.\ error $0.028$), no per-benchmark fitting. Right: predicted vs.\ empirical \emph{shared-error floor} (rate of unanimous agreement on a wrong answer; $\Delta$ is the per-benchmark error, mean abs.\ $0.009$), near zero on competition math and rising on GPQA and MMLU-Pro, where many-option multiple choice forces wrong answers to collide.}
\label{fig:law}
\end{figure}

With \emph{no per-benchmark fitting}, the parameter-free law tracks consensus behavior closely: across all seven benchmarks it matches empirical consensus accuracy to a mean absolute error of $0.028$ and the shared-error floor to $0.009$ (Table~\ref{tab:law}, Figure~\ref{fig:law}), the inputs $(a,\rho,s)$ measured directly on each and nothing fit to the consensus behavior. Nor is the fit an artifact of using each benchmark's own statistics: in a leave-one-benchmark-out test, predicting a held-out benchmark from the wrong-answer profile \emph{averaged over the other six} gives the same error (MAE $0.024$), so the structure transfers across benchmarks and is not tuned to any one of them (Appendix~\ref{app:closedform}).

\begin{table}[t]
\caption{\textbf{The parameter-free law predicts consensus behavior} across seven benchmarks (no per-benchmark fitting: $(a,\rho,s)$ measured, predictions in closed form). Mean absolute error is $0.028$ on consensus accuracy and $0.009$ on the shared-error floor. The floor is near zero on competition math, where wrong answers scatter, and largest on GPQA science ($0.030$, shared misconceptions) and MMLU-Pro ($0.143$, where ten-option multiple choice forces wrong answers to collide).}
\label{tab:law}
\centering\small
\begin{tabular}{@{}lcc@{}}
\toprule
\textbf{Benchmark} & \textbf{Consensus acc.\ (empirical / predicted)} & \textbf{Shared-error floor (empirical / predicted)} \\
\midrule
GSM8K          & $0.973 / 0.975$ & $0.014 / 0.010$ \\
MATH-500       & $0.956 / 0.959$ & $0.004 / 0.006$ \\
OlympiadBench  & $0.734 / 0.759$ & $0.040 / 0.022$ \\
AIME-2024      & $0.700 / 0.756$ & $0.000 / 0.011$ \\
AIME-2025      & $0.433 / 0.531$ & $0.000 / 0.016$ \\
GPQA           & $0.369 / 0.381$ & $\mathbf{0.030} / 0.030$ \\
MMLU-Pro       & $0.458 / 0.461$ & $\mathbf{0.143} / 0.155$ \\
\bottomrule
\end{tabular}
\end{table}

The law's chief practical consequence is the \emph{shared-error floor}: the rate at which the whole panel agrees on the same wrong answer, an error that is invisible to any agreement signal because the panel is confidently unanimous. We measure it near zero on competition math, where wrong answers scatter, but $0.030$ on GPQA, where independently trained models share graduate-science misconceptions. This both bounds the method and is a measurement about current models: their blind spots are largely idiosyncratic on math and partly shared on science. Inspecting the GPQA shared errors confirms the mechanism the law posits: the unanimous-but-wrong cases are not random coincidences but a \emph{shared convention or heuristic} that every model applies, e.g.\ reporting a thermochemical quantity in the textbook-default unit, or associating an absorption line with its most common astrophysical tracer (Appendix~\ref{app:shared}). The accuracy prediction is tighter on saturated sets and looser on the hardest (AIME-2024/2025), where the single-difficulty latent is an approximation. These residuals are reported in Table~\ref{tab:law}; per-benchmark fitting would have absorbed them.

\begin{figure}[t]
\centering
\includegraphics[width=\linewidth]{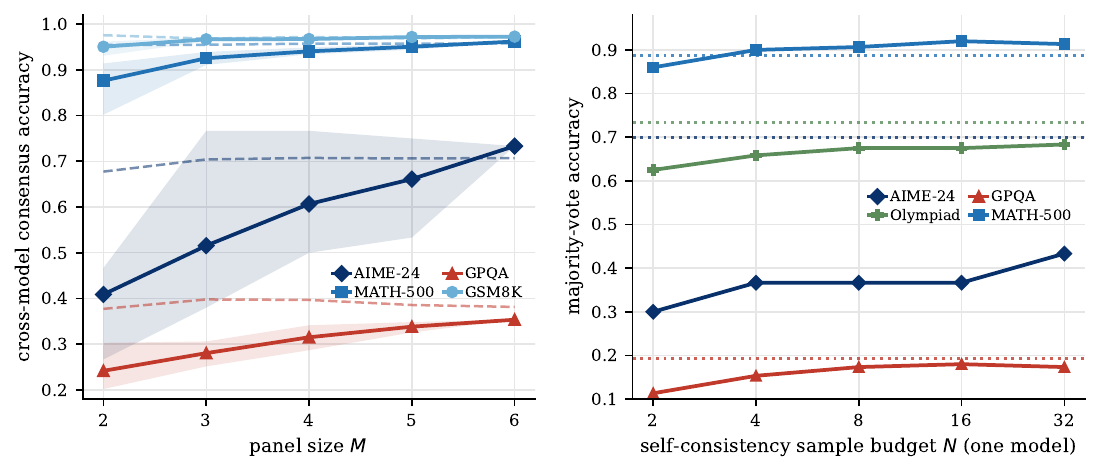}
\caption{\textbf{More models beats more samples.} \emph{Left:} consensus accuracy scales with panel size $M$: empirical cross-model accuracy (solid, averaged over all $M$-subsets of the pool; shaded band is their $10$--$90$th percentile) rises monotonically, and the parameter-free law (dashed) tracks it with no refitting. The gain is steepest where headroom exists (AIME-24) and flat where saturated (GSM8K). \emph{Right:} decorrelation beats more samples: self-consistency (solid) plateaus by $N{\approx}8$, while a four-model cross-model panel (dotted) exceeds $N{=}32$ self-consistency on AIME-24, OlympiadBench, and GPQA, and near-ties it on saturated MATH-500.}
\label{fig:scaling}
\end{figure}

\paragraph{The law forecasts the value of adding a model.} A natural question is whether enlarging the panel helps. Figure~\ref{fig:scaling} (left) sweeps panel size $M{=}2\!\to\!6$ (averaging over all $M$-subsets of the pool): consensus accuracy rises monotonically, most steeply where there is headroom (AIME-2024 $0.41\!\to\!0.73$) and gently where saturated (GSM8K $0.96\!\to\!0.99$), and the law's prediction tracks the empirical curve at every $M$ with no refitting, over-predicting only on the hardest sets as above.

\paragraph{Agreement is a usable abstention dial.} Beyond a single accuracy number, the law predicts the full \emph{selective-prediction curve} (accuracy as a function of the fraction of problems answered), which it tracks across benchmarks with no per-benchmark fitting (Appendix~\ref{app:selective}). This curve translates directly into an operating point a practitioner can read off (Table~\ref{tab:operating}). Answering only when the panel is \emph{unanimous} sharply raises accuracy: on MATH-500 a unanimous four-model panel is correct $99.5\%$ of the time while still answering $85\%$ of problems, and on AIME-2024/2025 the unanimous subset is $100\%$ correct. Relaxing the threshold to a three-quarters majority trades slightly lower accuracy for substantially higher coverage (MATH-500 $98.5\%$ at $94\%$ coverage). The two science/knowledge benchmarks behave as the shared-error floor predicts: unanimous-panel accuracy saturates at $0.769$ (GPQA) and $0.702$ (MMLU-Pro) instead of approaching one, because beyond that ceiling the panel is unanimous \emph{and} wrong, so no agreement threshold can recover it. Agreement thus serves as a calibrated abstention control whose maximum selective accuracy is fixed in advance by the floor.

\subsection{Cross-model consensus outperforms trained reward models}
\label{sec:prm}
Compared against four trained verifiers spanning the dominant paradigms (two discriminative process reward models, Qwen2.5-Math-PRM 7B and 72B~\citep{zhang2025prmlessons}; an outcome reward model, AceMath-72B-RM~\citep{liu2025acemath}; and a generative/reasoning verifier, ThinkPRM-14B~\citep{khalifa2025thinkprm}), the LLM-jury is the strongest non-oracle selector on all four benchmarks, at no training cost (Table~\ref{tab:prm}). On MATH-500, the trained verifiers' specialty, the jury ($0.966$) significantly exceeds the two discriminative PRMs and the generative verifier ($0.928$--$0.946$, paired bootstrap $p{\le}0.014$) and marginally exceeds the strongest trained verifier, the outcome RM AceMath-72B ($0.966$ vs.\ $0.956$, not significant at $p{=}0.142$). Out of domain on GPQA it is again the top selector ($0.359$ vs.\ $0.268$--$0.323$), beating the two math-specialist reward models and the generative verifier significantly ($p{\le}0.025$) and edging the $72$B discriminative PRM ($0.359$ vs.\ $0.323$, within noise). On the small-sample AIME sets it is the strongest selector or tied for it, reaching the pool's oracle on AIME-2024 ($0.567$, matching AceMath-72B) and leading on AIME-2025 ($0.400$). A label-free signal thus matches or exceeds state-of-the-art trained verifiers across the discriminative, outcome, and generative paradigms: it equals the best of them inside their own training domain and is the top selector out of it, at no training cost, because the trained verifier does not transfer whereas cross-model agreement is domain-agnostic. The advantage is clearest on heterogeneous tasks with no in-domain reward data, where a verifier is most needed and trained verifiers are least available.

The comparison is controlled: every selector operates on the same $N{=}12$ candidate pool of Table~\ref{tab:verifier}, so the head-to-head reflects the scoring signal alone, even though the trained verifiers read each candidate's full reasoning trace while the jury sees only final answers. The jury's advantage is not an artifact of a broken trained verifier: each is \emph{self-check gated}, its scores trusted only when they separate correct from incorrect candidates at AUROC $\ge 0.6$ and otherwise falling back to self-consistency, and on MATH-500 all four clear this gate ($0.74$--$0.86$) yet the jury still matches or exceeds them. Only out of domain does a verifier fall below it (ThinkPRM $0.57$ on GPQA).

\begin{table}[t]
\caption{\textbf{Cross-model consensus vs.\ four trained verifiers} as Best-of-$N$ selectors on the identical $N{=}12$ candidate pool of Table~\ref{tab:verifier}. The free jury is the strongest non-oracle selector on all four benchmarks, at no training cost: it beats three verifiers on MATH-500 ($p{\le}0.014$), edges the strongest (AceMath-72B, $0.966$ vs.\ $0.956$, $p{=}0.142$), and leads out-of-domain on GPQA. Per-benchmark significance is in the text (\S\ref{sec:prm}), and verifiers are self-check gated at AUROC $\ge 0.6$.}
\label{tab:prm}
\centering\small
\setlength{\tabcolsep}{5pt}
\begin{tabular}{@{}lcccc@{}}
\toprule
\textbf{Selector} & \textbf{MATH-500} & \textbf{AIME-24} & \textbf{AIME-25} & \textbf{GPQA} \\
& \footnotesize$n{=}500$ & \footnotesize$n{=}30$ & \footnotesize$n{=}30$ & \footnotesize$n{=}198$ \\
\midrule
Self-consistency            & $95.2$ & $36.7$ & $33.3$ & $33.3$ \\
Qwen2.5-Math-PRM-7B (disc.) & $92.8$ & $33.3$ & $33.3$ & $29.3$ \\
Qwen2.5-Math-PRM-72B (disc.)& $94.6$ & $50.0$ & $33.3$ & $32.3$ \\
AceMath-72B-RM (outcome)    & $95.6$ & $\mathbf{56.7}$ & $36.7$ & $30.3$ \\
ThinkPRM-14B (generative)   & $93.6$ & $33.3$ & $36.7$ & $26.8$ \\
\rowcolor{rowhl}
Cross-model consensus       & $\mathbf{96.6}$ & $\mathbf{56.7}$ & $\mathbf{40.0}$ & $\mathbf{35.9}$ \\
\midrule
Oracle (upper bound)        & $98.6$ & $56.7$ & $43.3$ & $53.0$ \\
\bottomrule
\end{tabular}
\end{table}

\paragraph{On cost.} The jury spends $M$ full generations where a PRM needs a single forward pass, but this cost is largely \emph{reuse}: the panel members are the strong models a practitioner already queries, so their solutions serve as both candidates and verifier with no reward data or separate scoring model to host. It can moreover be \emph{reduced} by an agreement-gated cascade: run a cheap two-model panel on every problem and escalate to the full four-model panel only on disagreement. Because disagreement concentrates on the hard problems, this recovers the full-panel accuracy at close to two-model cost (Figure~\ref{fig:cascade}; MATH-500 $0.956$ at $2.2$ calls/problem, AIME-2024 $0.700$ at $3.2$, escalation self-scaling from $3\%$ on GSM8K to $60$--$70\%$ on AIME/GPQA), and at matched average cost beats self-consistency by wide margins (AIME-2024 $0.700$ vs.\ $0.367$; Appendix~\ref{app:cascade}). Where in-domain reward data and a hosted PRM already exist, a trained verifier remains economical. Absent that data, the jury is both stronger and immediately deployable.

\paragraph{The signal generalizes beyond answer-matched selection.} Cross-model agreement is not tied to matching answer strings. On HumanEval+~\citep{liu2023evalplus}, where math PRMs are inapplicable by construction, we define consensus \emph{behaviorally} (two programs agree if they produce identical outputs on a shared battery of inputs) and find it a strong reliability signal: agreement predicts correctness at AUROC $0.748$, and programs in a unanimous behavioral class pass held-out tests $95.2\%$ of the time versus $52.9\%$ when the panel splits (Appendix~\ref{app:code}). The signal also trains models: as a label-free reward for rejection-sampling fine-tuning, peer consensus recovers most of the gain that \emph{gold} labels provide, at both $1.5$B and $7$B policy scales (Appendix~\ref{app:rft}).

\subsection{Decorrelation is the mechanism}
\label{sec:mechanism}
The law attributes the cross-model advantage to error decorrelation, and we confirm this directly: pairwise error correlation falls in the order within-model (self-consistency, $\bar\rho{=}0.68$) $>$ same-family ($0.52$) $>$ cross-family ($0.47$), exactly the ordering in which each construction improves consensus accuracy. Holding panel size and member accuracy fixed and varying only whether members are distinct, a panel of distinct models beats one model resampled as often ($+12.5$ to $+16.7$ on AIME-2024), a gain neither sample count nor raw strength can explain; a subset sweep, a leave-one-member-out ablation, and the accuracy/calibration split are in Appendix~\ref{app:mechanism}.

\paragraph{Decorrelation beats more samples.} Were the cross-model advantage merely ``more diversity is good,'' sampling the single generator more times would recover it. Sweeping the self-consistency budget to $N{=}32$ (Figure~\ref{fig:scaling}, right) shows otherwise, with single-model accuracy plateauing by $N{\approx}8$. The gap is largest on AIME-2024, where self-consistency saturates at $43.3\%$ ($N{=}32$) while a four-model panel reaches $70.0\%$ with $4$ generations, an $8\times$ smaller budget yet $+26.7$ points. The panel also leads at every $N$ on OlympiadBench and GPQA, and matches self-consistency only on saturated MATH-500 (both near ceiling). Resampling one model cannot escape that model's correlated errors at any $N$, whereas a decorrelated panel can. On problems with headroom, the budget is therefore better spent on distinct models than on additional samples of one.

\section{Conclusion}
\label{sec:conclusion}
We characterize the LLM-jury as a training-free verification primitive. It is the strongest selector against trained reward models both in and out of their training domain. A closed-form parameter-free law predicts its accuracy and its ceiling from measured panel statistics, and that law identifies the single failure mode, a shared misconception, which we measure and bound. The method requires neither reward training nor labels: a panel of three to four cross-family models, with the majority verdict as the answer and the agreement fraction as a calibrated abstention signal, recovers the full selection gain, and a split verdict routes additional compute through the agreement-gated cascade. To the extent that the selector, more than the generator, bounds test-time scaling, a decorrelated panel is a strong default that uses only the models a practitioner already has. Its remaining ceiling, the shared-error floor, is not a property of the method but of how alike current models are, and the law quantifies the value of reducing it by training models that fail differently.

\paragraph{Limitations.} The LLM-jury inherits a hard ceiling at the shared-error floor: where independently trained models share a misconception (measurably, parts of GPQA), unanimous agreement is confidently wrong and no agreement-based verifier can help, and selection likewise cannot exceed the pool's oracle. As a \emph{voting} rule it does not beat the single strongest panel member when panel strength is highly unequal, so its value lies in the \emph{label-free} reliability signal (one rarely knows the best member a priori) and the calibrated abstention it provides, beyond aggregation alone. It also requires more than one model, which a single-model deployment cannot supply.

\section*{Reproducibility Statement}
The LLM-jury is implemented entirely through prompting and majority aggregation, so no fine-tuning is required for the verifier results. The selector definitions (self-consistency, cross-model consensus, single-model LLM-verifier, oracle) and the agreement/equivalence rules are specified in \S\ref{sec:method} and Appendix~\ref{app:setup}. The parameter-free law, its measured inputs $(a,\rho,s)$, its closed form, and the validation against simulation are given in \S\ref{sec:law} and Appendix~\ref{app:closedform}. All seven benchmarks (AIME-2024/2025, MATH-500, OlympiadBench, GSM8K, GPQA, MMLU-Pro) and HumanEval+ are public, with splits and per-experiment problem counts in \S\ref{sec:experiments}. Grading uses \texttt{math\_verify} for math, unit-test execution for code, and exact match for multiple choice, applied identically to every method. Decoding settings, panel composition, the trained-PRM self-check gating, the bootstrap protocol, and the rejection-sampling fine-tuning recipe are detailed in Appendices~\ref{app:setup}--\ref{app:rft}. The trained-verifier comparison uses public open-weight checkpoints served locally (Qwen2.5-Math-PRM 7B/72B, AceMath-72B-RM, ThinkPRM-14B). The central decorrelation effect, the predictive law, and the verifier advantage all reproduce with a fully open-weight panel (Qwen3-235B, DeepSeek-V3.2, Kimi-K2.5; Appendix~\ref{app:gen2}, Table~\ref{tab:openpanel}), so the protocol does not depend on any proprietary model, though as with any hosted model, exact outputs may shift across provider-side updates.

\bibliography{nliu}

@inproceedings{wei2022chain,
  title={Chain-of-Thought Prompting Elicits Reasoning in Large Language Models},
  author={Wei, Jason and Wang, Xuezhi and Schuurmans, Dale and Bosma, Maarten and Ichter, Brian and Xia, Fei and Chi, Ed and Le, Quoc and Zhou, Denny},
  booktitle={Advances in Neural Information Processing Systems},
  year={2022}
}

@inproceedings{wang2023selfconsistency,
  title={Self-Consistency Improves Chain of Thought Reasoning in Language Models},
  author={Wang, Xuezhi and Wei, Jason and Schuurmans, Dale and Le, Quoc and Chi, Ed and Narang, Sharan and Chowdhery, Aakanksha and Zhou, Denny},
  booktitle={International Conference on Learning Representations},
  year={2023}
}

@article{lightman2023verify,
  title={Let's Verify Step by Step},
  author={Lightman, Hunter and Kosaraju, Vineet and Burda, Yuri and Edwards, Harri and Baker, Bowen and Lee, Teddy and Leike, Jan and Schulman, John and Sutskever, Ilya and Cobbe, Karl},
  journal={arXiv preprint arXiv:2305.20050},
  year={2023}
}

@inproceedings{snell2025scaling,
  title={Scaling {LLM} Test-Time Compute Optimally can be More Effective than Scaling Model Parameters},
  author={Snell, Charlie and Lee, Jaehoon and Xu, Kelvin and Kumar, Aviral},
  booktitle={International Conference on Learning Representations},
  year={2025}
}

@article{muennighoff2025s1,
  title={s1: Simple Test-Time Scaling},
  author={Muennighoff, Niklas and Yang, Zitong and Shi, Weijia and Li, Xiang Lisa and Fei-Fei, Li and Hajishirzi, Hannaneh and Zettlemoyer, Luke and Liang, Percy and Cand{\`e}s, Emmanuel and Hashimoto, Tatsunori},
  journal={arXiv preprint arXiv:2501.19393},
  year={2025}
}

@article{cobbe2021gsm8k,
  title={Training Verifiers to Solve Math Word Problems},
  author={Cobbe, Karl and Kosaraju, Vineet and Bavarian, Mohammad and Chen, Mark and Jun, Heewoo and Kaiser, Lukasz and Plappert, Matthias and Tworek, Jerry and Hilton, Jacob and Nakano, Reiichiro and Hesse, Christopher and Schulman, John},
  journal={arXiv preprint arXiv:2110.14168},
  year={2021}
}

@inproceedings{hendrycks2021math,
  title={Measuring Mathematical Problem Solving With the {MATH} Dataset},
  author={Hendrycks, Dan and Burns, Collin and Kadavath, Saurav and Arora, Akul and Basart, Steven and Tang, Eric and Song, Dawn and Steinhardt, Jacob},
  booktitle={Advances in Neural Information Processing Systems},
  year={2021}
}

@article{rein2024gpqa,
  title={{GPQA}: A Graduate-Level {Google-Proof} Q\&A Benchmark},
  author={Rein, David and Hou, Betty Li and Stickland, Asa Cooper and Petty, Jackson and Pang, Richard Yuanzhe and Dirani, Julien and Michael, Julian and Bowman, Samuel R.},
  journal={arXiv preprint arXiv:2311.12022},
  year={2024}
}

@misc{aime2024,
  title={American Invitational Mathematics Examination ({AIME}) 2024--2025},
  author={{Mathematical Association of America}},
  year={2025}
}

@article{huang2023selfcorrection,
  title={Large Language Models Cannot Self-Correct Reasoning Yet},
  author={Huang, Jie and Chen, Xinyun and Mishra, Swaroop and Zheng, Huaixiu Steven and Yu, Adams Wei and Song, Xinying and Zhou, Denny},
  journal={arXiv preprint arXiv:2310.01798},
  year={2023}
}

@inproceedings{du2024debate,
  title={Improving Factuality and Reasoning in Language Models through Multiagent Debate},
  author={Du, Yilun and Li, Shuang and Torralba, Antonio and Tenenbaum, Joshua B. and Mordatch, Igor},
  booktitle={International Conference on Machine Learning},
  year={2024}
}

@article{wang2024moa,
  title={Mixture-of-Agents Enhances Large Language Model Capabilities},
  author={Wang, Junlin and Wang, Jue and Athiwaratkun, Ben and Zhang, Ce and Zou, James},
  journal={arXiv preprint arXiv:2406.04692},
  year={2024}
}

@inproceedings{jiang2023llmblender,
  title={{LLM-Blender}: Ensembling Large Language Models with Pairwise Ranking and Generative Fusion},
  author={Jiang, Dongfu and Ren, Xiang and Lin, Bill Yuchen},
  booktitle={Proceedings of the 61st Annual Meeting of the Association for Computational Linguistics (ACL)},
  year={2023}
}

@article{wang2024mathshepherd,
  title={Math-Shepherd: Verify and Reinforce {LLMs} Step-by-step without Human Annotations},
  author={Wang, Peiyi and Li, Lei and Shao, Zhihong and Xu, Runxin and Dai, Damai and Li, Yifei and Chen, Deli and Wu, Yu and Sui, Zhifang},
  journal={arXiv preprint arXiv:2312.08935},
  year={2024}
}

@article{zhang2025prmlessons,
  title={The Lessons of Developing Process Reward Models in Mathematical Reasoning},
  author={Zhang, Zhenru and Zheng, Chujie and Wu, Yangzhen and Zhang, Beichen and Lin, Runji and Yu, Bowen and Liu, Dayiheng and Zhou, Jingren and Lin, Junyang},
  journal={arXiv preprint arXiv:2501.07301},
  year={2025}
}

@article{liu2025acemath,
  title={{AceMath}: Advancing Frontier Math Reasoning with Post-Training and Reward Modeling},
  author={Liu, Zihan and Chen, Yang and Shoeybi, Mohammad and Catanzaro, Bryan and Ping, Wei},
  journal={arXiv preprint arXiv:2412.15084},
  year={2024}
}

@inproceedings{zhang2025genrm,
  title={Generative Verifiers: Reward Modeling as Next-Token Prediction},
  author={Zhang, Lunjun and Hosseini, Arian and Bansal, Hritik and Kazemi, Mehran and Kumar, Aviral and Agarwal, Rishabh},
  booktitle={International Conference on Learning Representations (ICLR)},
  year={2025}
}

@article{khalifa2025thinkprm,
  title={Process Reward Models That Think},
  author={Khalifa, Muhammad and Agarwal, Rishabh and Logeswaran, Lajanugen and Kim, Jaekyeom and Peng, Hao and Lee, Moontae and Lee, Honglak and Wang, Lu},
  journal={arXiv preprint arXiv:2504.16828},
  year={2025}
}

@article{brown2024monkeys,
  title={Large Language Monkeys: Scaling Inference Compute with Repeated Sampling},
  author={Brown, Bradley and Juravsky, Jordan and Ehrlich, Ryan and Clark, Ronald and Le, Quoc V. and R{\'e}, Christopher and Mirhoseini, Azalia},
  journal={arXiv preprint arXiv:2407.21787},
  year={2024}
}

@article{zhao2025sample,
  title={Sample, Scrutinize and Scale: Effective Inference-Time Search by Scaling Verification},
  author={Zhao, Eric and Awasthi, Pranjal and Gollapudi, Sreenivas},
  journal={arXiv preprint arXiv:2502.01839},
  year={2025}
}

@inproceedings{wang2024mmlupro,
  title={{MMLU-Pro}: A More Robust and Challenging Multi-Task Language Understanding Benchmark},
  author={Wang, Yubo and Ma, Xueguang and Zhang, Ge and others},
  booktitle={Advances in Neural Information Processing Systems (NeurIPS) Datasets and Benchmarks Track},
  year={2024}
}

@inproceedings{he2024olympiadbench,
  title={{OlympiadBench}: A Challenging Benchmark for Promoting {AGI} with Olympiad-Level Bilingual Multimodal Scientific Problems},
  author={He, Chaoqun and Luo, Renjie and Bai, Yuzhuo and others},
  booktitle={Proceedings of the 62nd Annual Meeting of the Association for Computational Linguistics (ACL)},
  year={2024}
}

@inproceedings{liu2023evalplus,
  title={Is Your Code Generated by {ChatGPT} Really Correct? Rigorous Evaluation of Large Language Models for Code Generation},
  author={Liu, Jiawei and Xia, Chunqiu Steven and Wang, Yuyao and Zhang, Lingming},
  booktitle={Advances in Neural Information Processing Systems (NeurIPS)},
  year={2023}
}

@inproceedings{zheng2023judge,
  title={Judging {LLM-as-a-Judge} with {MT-Bench} and {Chatbot} {Arena}},
  author={Zheng, Lianmin and Chiang, Wei-Lin and Sheng, Ying and Zhuang, Siyuan and Wu, Zhanghao and Zhuang, Yonghao and Lin, Zi and Li, Zhuohan and Li, Dacheng and Xing, Eric P. and Zhang, Hao and Gonzalez, Joseph E. and Stoica, Ion},
  booktitle={Advances in Neural Information Processing Systems (NeurIPS) Datasets and Benchmarks Track},
  year={2023}
}

@article{verga2024jury,
  title={Replacing Judges with Juries: Evaluating {LLM} Generations with a Panel of Diverse Models},
  author={Verga, Pat and Hofstatter, Sebastian and Althammer, Sophia and Su, Yixuan and Piktus, Aleksandra and Arkhangorodsky, Arkady and Xu, Minjie and White, Naomi and Lewis, Patrick},
  journal={arXiv preprint arXiv:2404.18796},
  year={2024}
}

@article{chen2023universalsc,
  title={Universal Self-Consistency for Large Language Model Generation},
  author={Chen, Xinyun and Aksitov, Renat and Alon, Uri and Ren, Jie and Xiao, Kefan and Yin, Pengcheng and Prakash, Sushant and Sutton, Charles and Wang, Xuezhi and Zhou, Denny},
  journal={arXiv preprint arXiv:2311.17311},
  year={2023}
}

@article{kadavath2022know,
  title={Language Models (Mostly) Know What They Know},
  author={Kadavath, Saurav and Conerly, Tom and Askell, Amanda and Henighan, Tom and Drain, Dawn and Perez, Ethan and Schiefer, Nicholas and Hatfield-Dodds, Zac and DasSarma, Nova and Tran-Johnson, Eli and others},
  journal={arXiv preprint arXiv:2207.05221},
  year={2022}
}

@inproceedings{tian2023justask,
  title={Just Ask for Calibration: Strategies for Eliciting Calibrated Confidence Scores from Language Models Fine-Tuned with Human Feedback},
  author={Tian, Katherine and Mitchell, Eric and Zhou, Allan and Sharma, Archit and Rafailov, Rafael and Yao, Huaxiu and Finn, Chelsea and Manning, Christopher D.},
  booktitle={Proceedings of the 2023 Conference on Empirical Methods in Natural Language Processing (EMNLP)},
  year={2023}
}

@article{li2024moreagents,
  title={More Agents Is All You Need},
  author={Li, Junyou and Zhang, Qin and Yu, Yangbin and Fu, Qiang and Ye, Deheng},
  journal={Transactions on Machine Learning Research (TMLR)},
  year={2024}
}

@article{yue2025rlvr,
  title={Does Reinforcement Learning Really Incentivize Reasoning Capacity in {LLMs} Beyond the Base Model?},
  author={Yue, Yang and Chen, Zhiqi and Lu, Rui and Zhao, Andrew and Wang, Zhaokai and Yue, Yang and Song, Shiji and Huang, Gao},
  journal={Advances in Neural Information Processing Systems (NeurIPS)},
  year={2025}
}

@article{liu2025compute,
  title={Can {1B} {LLM} Surpass {405B} {LLM}? Rethinking Compute-Optimal Test-Time Scaling},
  author={Liu, Runze and Gao, Junqi and Zhao, Jian and Zhang, Kaiyan and Li, Xiu and Qi, Biqing and Ouyang, Wanli and Zhou, Bowen},
  journal={arXiv preprint arXiv:2502.06703},
  year={2025}
}

@inproceedings{krogh1994ensemble,
  title={Neural Network Ensembles, Cross Validation, and Active Learning},
  author={Krogh, Anders and Vedelsby, Jesper},
  booktitle={Advances in Neural Information Processing Systems (NeurIPS)},
  year={1994}
}

@article{dietterich2000ensemble,
  title={Ensemble Methods in Machine Learning},
  author={Dietterich, Thomas G.},
  journal={Multiple Classifier Systems},
  pages={1--15},
  year={2000},
  publisher={Springer}
}

@article{kuncheva2003diversity,
  title={Measures of Diversity in Classifier Ensembles and their Relationship with the Ensemble Accuracy},
  author={Kuncheva, Ludmila I. and Whitaker, Christopher J.},
  journal={Machine Learning},
  volume={51},
  number={2},
  pages={181--207},
  year={2003},
  publisher={Springer}
}

@article{lakshminarayanan2017deepensembles,
  title={Simple and Scalable Predictive Uncertainty Estimation using Deep Ensembles},
  author={Lakshminarayanan, Balaji and Pritzel, Alexander and Blundell, Charles},
  journal={Advances in Neural Information Processing Systems (NeurIPS)},
  year={2017}
}

@article{zhang2025hiddenstates,
  title={Reasoning Models Know When They're Right: Probing Hidden States for Self-Verification},
  author={Zhang, Anqi and Chen, Yulin and Pan, Jane and Zhao, Chen and Panda, Aurojit and Li, Jinyang and He, He},
  journal={arXiv preprint arXiv:2504.05419},
  year={2025}
}

@article{wang2024rethinking,
  title={Rethinking the Bounds of {LLM} Reasoning: Are Multi-Agent Discussions the Key?},
  author={Wang, Qineng and Wang, Zihao and Su, Ying and Tong, Hanghang and Song, Yangqiu},
  journal={arXiv preprint arXiv:2402.18272},
  year={2024}
}

@article{breiman2001random,
  title={Random Forests},
  author={Breiman, Leo},
  journal={Machine Learning},
  volume={45},
  number={1},
  pages={5--32},
  year={2001},
  publisher={Springer}
}
\bibliographystyle{iclr2026_conference}

\appendix

\section{Experimental Setup}
\label{app:setup}

\paragraph{Models and panels.} The cross-family pool comprises Qwen3-235B, DeepSeek-V3.2, Claude Sonnet~4.6, and Kimi-K2.5, accessed through a hosted inference API. The same-family control is the Qwen ladder (Qwen3-32B, Qwen3-Next-80B, Qwen3-235B). Unless noted, the Best-of-$N$ generator is Qwen3-235B and the cross-model panel for selection is \{DeepSeek-V3.2, Claude Sonnet~4.6, Kimi-K2.5\} plus the generator. Candidate pools use temperature $0.8$, and single answers use greedy decoding. Trained-verifier and rejection-sampling fine-tuning (RFT) experiments run open-weight models locally on a single multi-GPU node: the verifier bakeoff scores Qwen2.5-Math-PRM-7B/72B (discriminative; per-step $P(\text{good})$ read at each step separator and aggregated to a candidate score by the minimum, the standard choice for these models), AceMath-72B-RM (outcome; the single scalar from its reward head over the full solution), and ThinkPRM-14B (generative; served with \texttt{vllm}, scored from its \texttt{\textbackslash boxed\{correct/incorrect\}} step verdicts), and the RFT policies are Qwen2.5-7B/1.5B-Instruct. The frontier-panel ablation (\S\ref{app:gen2}) draws the selection panel from Claude Opus~4.8, Kimi-K2-Thinking, and Claude Sonnet~4.6.

\paragraph{Grading.} Answers are extracted and compared with the same harness across all methods: symbolic/numeric equivalence via \texttt{math\_verify} and a boxed-answer parser for math (MATH-500, AIME, OlympiadBench, GSM8K), and letter-choice match for multiple choice (GPQA, MMLU-Pro). The identical grader is used to score every selector, so comparisons are not confounded by extraction differences. We also re-grade all RFT generations with this harness instead of any in-loop grader (\S\ref{app:rft}).

\paragraph{Statistics.} Significance uses one-sided paired bootstrap $p$-values, resampling the per-problem win/loss difference between two selectors on the same problems ($10$K resamples). AIME-2024 and AIME-2025 are pooled where noted for power ($n{=}60$). Table~\ref{tab:ci} reports the accompanying $95\%$ bootstrap confidence interval on every cell of the headline comparison (Table~\ref{tab:verifier}). The marginal intervals are wide on the small sets (half-width $\pm16.7$ at $n{=}30$, versus $\pm1.5$ at $n{=}500$), and the cross-model and self-consistency intervals overlap on several benchmarks even where the difference is significant: this is expected, because the paired test cancels the shared per-problem difficulty and is therefore the correct instrument on small samples, whereas non-overlap of marginal intervals is a needlessly conservative and less powerful criterion. We report both so the reader can see the raw dispersion and the paired comparison side by side.

\begin{table}[ht]
\caption{\textbf{Per-cell $95\%$ bootstrap confidence intervals} for the headline selection comparison (Table~\ref{tab:verifier}; $10$K resamples, subscripts give the $[2.5,97.5]$ percentile interval in points). Interval half-width scales with sample size, from $\pm16.7$ at $n{=}30$ to $\pm1.5$ at $n{=}500$. The significance of each cross-model advantage is established by the paired bootstrap in the text; non-overlap of these marginal intervals is not the criterion.}
\label{tab:ci}
\centering\footnotesize
\setlength{\tabcolsep}{3pt}
\begin{tabular}{@{}lccccccc@{}}
\toprule
\textbf{Selector} & \textbf{AIME-24} & \textbf{AIME-25} & \textbf{MATH} & \textbf{Olympiad} & \textbf{GPQA} & \textbf{MMLU-Pro} & \textbf{GSM8K} \\
& \scriptsize$n{=}30$ & \scriptsize$n{=}30$ & \scriptsize$n{=}500$ & \scriptsize$n{=}674$ & \scriptsize$n{=}198$ & \scriptsize$n{=}1000$ & \scriptsize$n{=}1319$ \\
\midrule
Self-consistency      & $36.7_{20\text{-}53}$ & $33.3_{17\text{-}50}$ & $95.2_{93\text{-}97}$ & $65.4_{62\text{-}69}$ & $33.3_{27\text{-}40}$ & $43.8_{41\text{-}47}$ & $96.7_{96\text{-}98}$ \\
LLM-verifier          & $30.0_{13\text{-}47}$ & $33.3_{17\text{-}50}$ & $93.2_{91\text{-}95}$ & $62.2_{59\text{-}66}$ & $26.3_{20\text{-}32}$ & $43.9_{41\text{-}47}$ & $95.9_{95\text{-}97}$ \\
\rowcolor{rowhl}
Cross-model consensus & $56.7_{40\text{-}73}$ & $40.0_{23\text{-}57}$ & $96.6_{95\text{-}98}$ & $72.6_{69\text{-}76}$ & $35.9_{29\text{-}42}$ & $46.2_{43\text{-}49}$ & $97.2_{96\text{-}98}$ \\
Oracle (upper bound)  & $56.7_{40\text{-}73}$ & $43.3_{27\text{-}60}$ & $98.6_{97\text{-}100}$ & $79.1_{76\text{-}82}$ & $53.0_{46\text{-}60}$ & $48.9_{46\text{-}52}$ & $97.9_{97\text{-}99}$ \\
\bottomrule
\end{tabular}
\end{table}

\paragraph{Selector definitions.} Self-consistency draws $N$ samples from one model (temperature $0.8$) and majority-votes. Cross-model consensus takes the panel-majority over $M{=}N$ independent models. The LLM-verifier prompts one model to score each candidate $0$--$10$ and selects the argmax. Oracle selects any correct candidate. Agreement classes use the grading equivalence above.

As a Best-of-$N$ selector the jury always returns a member of the candidate pool: it scores each \emph{candidate} by the number of independent panel members whose answer is equivalent to it, and returns the highest-scoring candidate. This resolves the edge case where the panel's modal answer is \emph{absent} from the pool: such an answer receives no support because no candidate matches it, and the rule falls back to the next-best-supported candidate, degrading gracefully to the self-consistency (pool-frequency) choice when the panel and pool share no answer. Ties are broken first by pool frequency, then by earliest sampled candidate, so the selection is deterministic and independent of hash order.

\paragraph{Prompts.} The jury is prompt-only, so we give the exact prompts verbatim. Every panel member solves each problem with the same instruction, which asks for numbered steps with an explicit named quantity per step (this step structure is what the trained PRMs later score, while the jury itself reads only the final answer):
\begin{promptbox}[Solver prompt (every panel member and generator)]
\begin{Verbatim}
[system] You are an expert problem solver. Solve the given problem
carefully and show your full reasoning. Be rigorous: every numerical
or logical result you state should follow from what came before.

[user] Problem: {problem}
Solve this problem step by step. Number each logical step. At each
step, make the intermediate QUANTITY you compute explicit as a
`name = value` pair at the end of the step, where the name says WHAT
the quantity is (in the problem's own terms, e.g. `discriminant`,
`num_valid_cases`). Format strictly as:
STEP 1: <reasoning> => <quantity_name> = <value>
...
FINAL ANSWER: <answer>
\end{Verbatim}
\end{promptbox}
The single-model LLM-verifier baseline scores each candidate with one call, taking the argmax score over the pool:
\begin{promptbox}[Single-model verifier prompt]
\begin{Verbatim}
[system] You are a careful judge scoring whether a candidate solution
to a problem is correct. You reason about its validity and output a
single numeric score.

[user] Problem: {problem}
Candidate solution: {solution}
Judge whether this solution's reasoning and final answer are correct.
Consider the method, arithmetic, and whether the conclusion follows.
End with exactly:
SCORE: <0-10, where 10 = certainly correct, 0 = certainly wrong>
\end{Verbatim}
\end{promptbox}
The verifier is scored at temperature $0$, and the score is parsed from the trailing \texttt{SCORE:} line. Both prompts are model-agnostic: every panel member and every generator receives the identical text, so no per-model prompt tuning enters the comparison.

\section{The Predictive Law: Details}
\label{app:law}
The latent-difficulty parameter $p$ is drawn from $\mathrm{Beta}(ak, (1{-}a)k)$ with $k$ chosen so the Bernoulli-error correlation between two models equals the measured $\rho$ (as $\rho{\to}0$, $k{\to}\infty$ and $p{\to}a$, the independent case). $(a,\rho,s)$ are estimated directly on each benchmark: $a$ as mean per-model accuracy, $\rho$ as the mean pairwise correlation of model error indicators, and $s$ as the fraction of wrong answers falling in the modal wrong class. These are descriptive panel statistics: no quantity is fit to the empirical consensus behavior the law predicts, so every benchmark is effectively held out in that sense (the leave-one-benchmark-out test of Appendix~\ref{app:closedform} checks the stronger form, that the wrong-answer structure transfers across benchmarks).

\paragraph{What the law requires, and what it does not.} A distinction is worth drawing, since the two uses of the method have different data requirements. \emph{Running} the LLM-jury as a selector needs no labels: it returns the modal answer class regardless of correctness, which is the sense in which the verifier is label-free. Labels enter only when we \emph{predict} the jury's behavior with the law: its inputs are defined through correctness ($a$ is accuracy, $\rho$ correlates error indicators, and $s$ is a wrong-answer mass), so they must be measured on a sample with ground truth. The law is thus \emph{parameter-free} (no quantity is fit to the prediction target) but calibrated from a labeled sample: characterizing a panel on a new domain requires a labeled calibration set from that domain. Two facts bound this cost. First, the estimate is stable at modest sample size: the input-uncertainty intervals of Table~\ref{tab:predci} are tight by $n{=}500$ (a $90\%$ consensus-accuracy interval of $\pm0.014$) and already usable at $n{=}198$, so a few hundred labeled problems suffice, and as labels shrink the law degrades gracefully, with the predictive interval widening smoothly. Second, the most data-hungry input transfers: the leave-one-benchmark-out test (Appendix~\ref{app:closedform}) predicts a held-out domain's behavior from the wrong-answer profile of \emph{other} domains at unchanged error, so the multi-bin profile $s$ need not be re-measured on the target domain. Only the two scalars $(a,\rho)$, which stabilize fastest, remain domain-local. The law therefore characterizes a panel in advance from a small labeled calibration set, while the selector it describes runs unlabeled.

\section{Closed-Form Law and Validation}
\label{app:closedform}
The predictions of \S\ref{sec:law} are exact functions of $(a,\rho,s,M)$ and do not rely on simulation. Marginalizing the latent difficulty $p\sim\mathrm{Beta}(ak,(1{-}a)k)$ ($k{=}1/\rho-1$), the number of correct members $C$ over a panel of $M$ follows a Beta-Binomial, $\Pr[C{=}c]=\binom{M}{c}\,B(c+ak,\,M-c+(1{-}a)k)/B(ak,(1{-}a)k)$, where $B$ is the Beta function.

\paragraph{Shared-error floor.} The panel is unanimously wrong iff all $M$ members err and all land on the shared attractor. Since the two events factor given $p$, $\mathbb{E}[(1-p)^M]=\prod_{j=0}^{M-1}\frac{(1-a)k+j}{k+j}$, giving Eq.~\ref{eq:floor}: $\mathrm{Floor}=s^M\,\mathbb{E}[(1-p)^M]$.

\paragraph{Consensus accuracy.} With idiosyncratic wrong answers distinct (singletons), the modal class is the correct one iff $C\ge 1$ and no attractor class exceeds $C$. Conditioning on $C{=}c$, the $M-c$ wrong members are multinomial over the attractor (prob.\ $s$) and idiosyncratic outcomes, so
\begin{equation}
\mathrm{Acc}=\sum_{c=1}^{M}\Pr[C{=}c]\,\Pr[\,A\le c\mid M-c\text{ wrong}\,],\qquad A\sim\mathrm{Binom}(M-c,\,s),
\end{equation}
with ties resolved to the correct class. The selective-prediction curve is the same sum restricted to problems whose agreement level $g$ exceeds a threshold. These closed forms (evaluated with each benchmark's measured wrong-answer profile) match Monte-Carlo simulation ($3{\times}10^5$ problems) to within $0.002$ on every benchmark (Table~\ref{tab:closedform}), confirming the closed forms are exact up to sampling error.

\begin{table}[ht]
\caption{\textbf{The closed-form law matches simulation} to within $0.002$ (consensus accuracy) and $0.001$ (floor) on every benchmark, confirming the closed forms (Eq.~\ref{eq:floor} and its accuracy counterpart, evaluated with each benchmark's measured wrong-answer profile) are exact.}
\label{tab:closedform}
\centering\small
\setlength{\tabcolsep}{5pt}
\begin{tabular}{@{}lcccc@{}}
\toprule
& \multicolumn{2}{c}{\textbf{Consensus acc.}} & \multicolumn{2}{c}{\textbf{Shared-error floor}} \\
\cmidrule(lr){2-3}\cmidrule(lr){4-5}
\textbf{Benchmark} & \textbf{closed-form} & \textbf{simulation} & \textbf{closed-form} & \textbf{simulation} \\
\midrule
GSM8K          & $0.975$ & $0.975$ & $0.010$ & $0.010$ \\
MATH-500       & $0.958$ & $0.959$ & $0.006$ & $0.006$ \\
AIME-2024      & $0.756$ & $0.754$ & $0.011$ & $0.011$ \\
AIME-2025      & $0.530$ & $0.529$ & $0.016$ & $0.017$ \\
OlympiadBench  & $0.758$ & $0.757$ & $0.022$ & $0.022$ \\
GPQA           & $0.382$ & $0.380$ & $0.031$ & $0.031$ \\
MMLU-Pro       & $0.461$ & $0.461$ & $0.155$ & $0.154$ \\
\bottomrule
\end{tabular}
\end{table}

\paragraph{Uncertainty in the measured inputs.} The law's inputs $(a,\rho,\text{masses})$ are themselves estimated on a finite sample, so on the small sets they carry sampling noise. We propagate this noise by bootstrapping the estimation set (resampling problems with replacement, re-measuring the inputs on each resample, and pushing them through the closed form) to obtain a predictive interval on the law's output (Table~\ref{tab:predci}). The interval is wide exactly where the sample is small (AIME, $n{=}30$: consensus-accuracy $90\%$ interval spanning $\approx0.17$) and tight where it is large (MATH-500, $n{=}500$: $\approx0.03$), as expected. In every case the empirical consensus accuracy and shared-error floor fall inside the law's predictive interval, so the point-prediction residuals reported in \S\ref{sec:theory} are within what estimation noise alone predicts. The law is not systematically biased, it is simply measured less precisely on the small sets.

\begin{table}[ht]
\caption{\textbf{Predictive intervals from input uncertainty.} Bootstrapping the estimation set ($2$K resamples, re-measuring $(a,\rho,\text{masses})$ on each and evaluating the closed form) gives a $90\%$ predictive interval on the law's output. The interval widens as sample size falls, and contains the empirical value on every benchmark, so the small-$n$ residuals are consistent with estimation noise and do not indicate model bias.}
\label{tab:predci}
\centering\small
\setlength{\tabcolsep}{6pt}
\begin{tabular}{@{}lccc@{}}
\toprule
\textbf{Benchmark} & \textbf{Empirical acc.} & \textbf{Predicted acc.\ ($90\%$ interval)} & \textbf{Predicted floor ($90\%$)} \\
\midrule
AIME-2024 ($n{=}30$)  & $0.700$ & $0.756\ [0.670,\,0.841]$ & $0.011\ [0.007,\,0.017]$ \\
AIME-2025 ($n{=}30$)  & $0.433$ & $0.528\ [0.401,\,0.657]$ & $0.016\ [0.011,\,0.024]$ \\
GPQA ($n{=}198$)      & $0.369$ & $0.380\ [0.333,\,0.429]$ & $0.031\ [0.025,\,0.038]$ \\
MATH-500 ($n{=}500$)  & $0.956$ & $0.958\ [0.944,\,0.971]$ & $0.006\ [0.004,\,0.010]$ \\
\bottomrule
\end{tabular}
\end{table}

\paragraph{Held-out transfer.} Because the law has no free parameters, every benchmark is already held out in the sense that nothing is fit to its consensus behavior. A stronger test asks whether the \emph{wrong-answer structure} transfers: in a leave-one-benchmark-out protocol we take the wrong-answer profile averaged over six benchmarks and predict the seventh's consensus accuracy from it plus the seventh's measured $(a,\rho)$ (still no quantity optimized against the consensus target, since only the profile's \emph{source} changes from the held-out set to the other six). This leaves the mean absolute error essentially unchanged ($0.028$ with each benchmark's own profile, $0.024$ with the held-out profile), so the structure the law relies on is shared across domains instead of being benchmark-specific.

\paragraph{Transfer across panels.} The law is validated in \S\ref{sec:theory} on the default cross-family panel. A stronger test asks whether the \emph{same} law, unchanged, predicts a different panel it is not tuned on. We apply it to the frontier panel of \S\ref{app:gen2} (Qwen3-235B plus Claude Opus~4.8, Kimi-K2-Thinking, and Claude Sonnet~4.6), measuring that panel's own $(a,\rho,\text{wrong-answer profile})$ and predicting its empirical consensus accuracy and shared-error floor with no other change. The prediction remains accurate: consensus-accuracy MAE is $0.036$ and floor MAE is $0.014$ across six benchmarks, comparable to the default panel and with the same signature (the only sizeable miss is the $30$-problem AIME-2024, where the law over-predicts $0.791$ vs.\ $0.900$ empirical, the hardest-set approximation noted in \S\ref{sec:theory}). The law therefore describes cross-model panels as a class and is not specific to one roster.

\paragraph{Multi-attractor refinement.} The single-attractor model assumes wrong answers either scatter or concentrate on one shared attractor. This holds on math and graduate science (GPQA), but a benchmark whose answer space \emph{forces} collisions (ten-option MMLU-Pro, where erring models frequently pick the same distractor by construction) violates it, and the single-attractor floor overshoots substantially ($0.41$ predicted vs.\ $0.14$ empirical). Replacing the scalar $s$ with a measured mass profile $(s_1,\dots,s_K)$ over the top-$K$ wrong-answer clusters (a wrong member lands on cluster $j$ with probability $s_j$) generalizes Eq.~\ref{eq:floor} to $\mathrm{Floor}=\big(\sum_j s_j^M\big)\mathbb{E}[(1-p)^M]$ and the accuracy sum analogously. With $K{=}6$ measured masses this cuts the floor mean absolute error from $0.046$ (scalar $s$) to $0.009$ across all seven benchmarks, and the consensus-accuracy MAE from $0.055$ to $0.028$ (Table~\ref{tab:multiattractor}): MMLU-Pro's floor drops to $0.155$ (vs.\ $0.143$ empirical) and OlympiadBench's to $0.022$ (vs.\ $0.040$). The main-text predictions (Table~\ref{tab:law}, Figure~\ref{fig:law}) use this measured profile. We retain the scalar-$s$ form of Equation~\ref{eq:floor} in the main text because its single parameter \emph{is} the shared-misconception rate and is the more interpretable object, and the two coincide wherever one misconception dominates (all benchmarks except the collision-prone multiple-choice sets).

Although we measure the profile with $K{=}6$ slots, the \emph{effective} $K$ is small: on every benchmark the wrong-answer mass concentrates on the first two-to-four attractors and the remaining slots are empty. The leading masses are $(0.94,0.06)$ on GSM8K and $(0.78,0.18,0.04)$ on MATH-500 (one dominant misconception, so the scalar-$s$ form suffices), versus the flatter $(0.49,0.25,0.17,0.09)$ on GPQA and $(0.75,0.21,0.04)$ on MMLU-Pro (several competing attractors, where the profile matters). The refinement therefore adds at most three effective numbers over the scalar $s$, and those numbers are not tuned to the prediction target: the leave-one-benchmark-out test above holds out an entire benchmark's profile and the accuracy MAE is unchanged ($0.024$), so the mass profile is a stable property of the answer space that transfers across benchmarks.

\begin{table}[ht]
\caption{\textbf{Single- vs.\ multi-attractor floor.} The single-attractor model (one scalar $s$) is accurate where wrong answers scatter or share one misconception, but overshoots on collision-prone MMLU-Pro. A measured top-$K$ mass profile fixes this, cutting floor MAE from $0.046$ to $0.009$.}
\label{tab:multiattractor}
\centering\small
\setlength{\tabcolsep}{6pt}
\begin{tabular}{@{}lccc@{}}
\toprule
\textbf{Benchmark} & \textbf{Empirical floor} & \textbf{Single-attractor} & \textbf{Multi-attractor} \\
\midrule
GSM8K          & $0.014$ & $0.009$ & $0.010$ \\
MATH-500       & $0.004$ & $0.003$ & $0.006$ \\
AIME-2024      & $0.000$ & $0.000$ & $0.011$ \\
AIME-2025      & $0.000$ & $0.005$ & $0.016$ \\
OlympiadBench  & $0.040$ & $0.005$ & $0.022$ \\
GPQA           & $0.030$ & $0.023$ & $0.030$ \\
MMLU-Pro       & $0.143$ & $0.411$ & $0.155$ \\
\midrule
\textbf{Mean abs.\ error} & --- & $0.046$ & $\mathbf{0.009}$ \\
\bottomrule
\end{tabular}
\end{table}

\section{The Shared-Error Floor: A Case Study}
\label{app:shared}
The law identifies a hard ceiling at the \emph{shared-error floor}: problems on which the entire panel is unanimous yet wrong, which no agreement-based verifier can flag. On GPQA the four-model cross-family panel is unanimously wrong on $6/198$ problems ($0.030$), versus $0/530$ on the pooled math sets. We inspect all six to ask whether the floor is random coincidence or a structured, shared blind spot, as the law's shared-misconception parameter $s$ assumes.

Two of the six are not genuine errors but grading artifacts where the unanimous answer is correct in a different surface form: a titration problem whose agreed answer (``pH $4.26$ at $25\%$ and $8.52$ at equivalence'') matches the gold (``$4.26$; $8.52$'') in prose form instead of the expected delimited form, and a multiple-select physics item where ``$1$, $2$, and $4$'' matches the gold ``$1,2,4$''. Excluding these, the true shared-error rate is $4/198\approx 0.020$, so the reported $0.030$ is a conservative upper bound.

The remaining four are real, and each stems from a \emph{shared convention or heuristic} instead of an idiosyncratic slip, exactly the structure $s$ models:
\begin{itemize}
\itemsep0.15em
\item \textbf{Enthalpy of formation (Chemistry).} All four models compute the bond-energy enthalpy and report $1900$~kJ/mol, but the gold answer is $11.44$~kJ/g, the same quantity converted to a per-gram basis. The models share the convention of reporting molar enthalpy.
\item \textbf{Absorption line at $2.1$~Gpc (Physics).} Given an absorption-line energy of $3.9\,\mu$eV, all four associate it with the $21$~cm hydrogen line and answer accordingly, but the gold answer identifies the cold atomic interstellar medium. The models share the standard line-to-tracer association.
\item \textbf{Synchrocyclotron revolutions (Physics).} All four set up the accelerating-phase counting the same way and converge on $2500$ revolutions, but the gold is $3536$. A shared modeling assumption produces the common wrong number; the errors do not scatter as arithmetic slips would.
\item \textbf{Fluorinated stereochemistry (Chemistry).} All four identify the difluoro product but omit the same stereochemical descriptors that the gold answer requires.
\end{itemize}
In every genuine case the models fail \emph{together and for the same reason}, confirming that the floor reflects the shared-misconception regime the law predicts and is not attributable to chance agreement. This is the precise sense in which cross-model consensus has a known limit: it is blind exactly where independently trained models have internalized the same convention, and our measurement localizes that regime to chemistry and physics (the six cases are split evenly between them, and none are in biology).

\section{Mechanism: Diversity Isolation and the Combinatorial Sweep}
\label{app:mechanism}

\paragraph{Pairwise error correlation.} Table~\ref{tab:rho} reports the mean pairwise error correlation $\rho$ of three panel constructions at matched size. Resampling one model leaves errors strongly correlated ($\bar\rho{=}0.68$), a same-family panel decorrelates them somewhat ($0.52$), and a cross-family panel most ($0.47$). By the law, lower $\rho$ raises consensus accuracy, and the ordering within-model $>$ same-family $>$ cross-family is precisely the ordering of how much each construction helps (\S\ref{sec:mechanism}).

\begin{table}[ht]
\caption{\textbf{Pairwise error correlation $\rho$ by panel construction.} Resampling one model keeps errors correlated ($\bar\rho{=}0.68$), while distinct models decorrelate them, most so across families ($0.47$). This is the variable the law identifies as decisive for consensus accuracy (\S\ref{sec:law}).}
\label{tab:rho}
\centering\small
\begin{tabular}{@{}lccccc|c@{}}
\toprule
\textbf{Panel ($\rho$, lower = more decorrelated)} & \textbf{GSM8K} & \textbf{MATH} & \textbf{GPQA} & \textbf{AIME-24} & \textbf{AIME-25} & \textbf{mean} \\
\midrule
within-model (self-consistency) & $0.70$ & $0.67$ & $0.60$ & $0.76$ & $0.67$ & $0.68$ \\
same-family (Qwen ladder)       & $0.45$ & $0.57$ & $0.43$ & $0.66$ & $0.51$ & $0.52$ \\
\rowcolor{rowhl}
cross-family                    & $0.48$ & $0.47$ & $0.46$ & $0.41$ & $0.56$ & $\mathbf{0.47}$ \\
\bottomrule
\end{tabular}
\end{table}

\paragraph{Matched-accuracy diversity isolation.} Drawing on the four cross-family models of \S\ref{app:setup}, we compare a within-model panel (each model sampled $k$ times, averaged over the four) against a cross-model panel ($k$-subsets of the four, one sample each) at matched member accuracy. On AIME-2024 the cross-model panel improves consensus accuracy by $+13.3$ ($k{=}2$), $+16.7$ ($k{=}3$), and $+12.5$ ($k{=}4$) over the within-model panel, yet its agreement-to-correctness AUROC is \emph{lower} at the same $k$ (by $0.10$ to $0.13$), the calibration/accuracy split discussed in \S\ref{sec:mechanism}. On the frontier-equal AIME-2025 and GPQA panels the accuracy gain all but vanishes (between $-0.8$ and $+3.6$ across $k$, straddling zero), consistent with little decorrelation left to exploit. To our knowledge this is the first measurement to separate the accuracy and calibration contributions of cross-model agreement relative to self-consistency: decorrelation is decisive for \emph{selecting} the correct answer, while within-model agreement remains the better confidence signal for \emph{ranking problems by reliability} (its agreement level more sharply separates the panel's correct decisions from its incorrect ones).

\paragraph{Combinatorial subset sweep.} Over all $\binom{M}{k}$ panels ($k{=}2..5$) drawn from the pool, we regress each panel's consensus gain over its best single member on member accuracy and pairwise error correlation. The diversity term is consistently significant: consensus gain correlates with lower error correlation at $r{=}{+}0.32$ to ${+}0.66$ across benchmarks, while calibration (agreement-to-correctness AUROC) tracks member accuracy. This is the empirical counterpart of the two-force account in \S\ref{sec:mech-theory}.

\paragraph{Leave-one-member-out.} Removing each member from the four-model cross-family panel in turn changes consensus accuracy little and is rarely the strongest member that matters: on AIME-2024 dropping the two Claude/Kimi members lowers accuracy ($0.70\!\to\!0.50$) while dropping Qwen or DeepSeek \emph{raises} it ($\to0.77$), and on the saturated sets all leave-one-member-out variants stay within $\pm1.5$ points of the full panel. The consensus signal is a property of the panel's collective decorrelation and does not derive from a single dominant model.

\section{Consensus as a Label-Free Training Reward}
\label{app:rft}
Beyond verification, cross-model consensus can serve as a label-free \emph{reward} to improve a model by rejection-sampling fine-tuning (RFT; \S\ref{app:setup}): sample $K$ solutions per training prompt, keep those whose answer matches a reward target, and fine-tune on the kept solutions. We compare three reward targets: \textbf{cross} (matches the cross-model peer consensus, label-free), \textbf{self} (matches the policy's own self-consistency majority, label-free), and \textbf{gold} (matches the ground-truth answer, an oracle that uses labels). The policy is Qwen2.5-1.5B-Instruct (chosen for genuine headroom; base MATH-500 $0.55$), trained on $6{,}000$ GSM8K prompts, and the peers are three open cross-family models (Phi-3.5-mini, DeepSeek-Math-7B, Mathstral-7B). All generations are re-graded with the harness of \S\ref{app:setup}.

\paragraph{Training details.} We sample $K{=}8$ candidates per prompt at temperature $0.8$, top-$p$ $0.95$, up to $1024$ tokens, and keep up to two matching solutions per prompt as supervised targets. Fine-tuning is LoRA (rank $32$, $\alpha{=}64$, dropout $0.05$) applied to all attention and MLP projections, trained for three epochs with AdamW at learning rate $1{\times}10^{-5}$, effective batch size $16$ (per-device $4$, gradient accumulation $4$), maximum sequence length $2048$, and \texttt{bfloat16}. Evaluation is greedy ($\text{temperature}{=}0$, up to $2048$ tokens). The $7$B run (Table~\ref{tab:rft7b}) uses the identical recipe and the same $6{,}000$ GSM8K prompts, changing only the policy to Qwen2.5-7B (base). We report a single training run per arm. The deltas are therefore point estimates without a variance estimate, so we interpret only the qualitative ordering (cross vs.\ self vs.\ gold) and treat sub-point differences as within run-to-run noise.

\begin{table}[ht]
\caption{Consensus as a label-free RFT reward (Qwen2.5-1.5B-Instruct; $\Delta$ accuracy points over the base policy, re-graded with the harness of \S\ref{app:setup}). On the in-distribution math benchmark the cross-model reward captures most of the oracle (gold) gain \emph{without labels} and clearly beats a self-consistency reward. The ordering cross $\approx$ gold $\gg$ self is the training-time analogue of the decorrelation mechanism (\S\ref{sec:mech-theory}). Out-of-domain GPQA and the saturated GSM8K training source move little for any reward.}
\label{tab:rft}
\centering\small
\begin{tabular}{@{}lccc@{}}
\toprule
\textbf{Reward (accuracy $\Delta$ over base)} & \textbf{MATH-500} & \textbf{GSM8K} & \textbf{GPQA (OOD)} \\
\midrule
self-consistency (label-free)   & $+1.5$ & $-2.3$ & $-0.7$ \\
\rowcolor{rowhl}
cross-model consensus (label-free) & $\mathbf{+4.5}$ & $-1.3$ & $-0.7$ \\
gold (oracle, uses labels)      & $+5.5$ & $+0.3$ & $+0.7$ \\
\bottomrule
\end{tabular}
\end{table}

Table~\ref{tab:rft} shows cross-model consensus is a usable label-free reward: on the in-distribution MATH-500 it lifts the policy by $+4.5$ points, most of the $+5.5$ gain from training on \emph{gold} labels, while a self-consistency reward yields only $+1.5$, because resampling the policy reinforces exactly the systematic errors a decorrelated panel does not share. The headroom of the policy matters: with a near-saturated policy (Qwen2.5-7B-Instruct, base MATH $0.76$) no reward, including gold, moved accuracy, indicating that the operative variable is policy headroom.

\paragraph{The label-free reward scales to a 7B policy.} To test whether this is a small-model artifact, we repeat the experiment at $4{\times}$ the policy scale, on Qwen2.5-7B (the \emph{base} model, base MATH-500 $0.505$, chosen so it has the headroom a saturated instruct model lacks), changing only the policy. The label-free consensus reward continues to track the oracle: trained on GSM8K, it transfers to MATH-500 for $+8.5$ points, $\approx 81\%$ of the $+10.5$ that \emph{gold} labels deliver, and improves the in-distribution GSM8K and out-of-domain GPQA as well (Table~\ref{tab:rft7b}). Consensus-as-reward is therefore not a property of weak policies: a free, label-free signal recovers most of the gain of ground-truth supervision at 7B. At this scale the cross-vs-self gap narrows (both lift MATH by $\approx 9$ points), consistent with the mechanism: a stronger policy's own samples are already reliable enough that self-consistency is a near-as-good target, whereas a weak policy (the 1.5B above) reinforces its own systematic errors and only a decorrelated panel escapes them. Cross still matches or exceeds the self-consistency reward on every benchmark. We re-grade all generations with the harness of \S\ref{app:setup}. On-policy RL and larger policies remain for future work.

\begin{table}[ht]
\caption{\textbf{Consensus-as-reward scales to a 7B policy} (Qwen2.5-7B base, trained on GSM8K; $\Delta$ accuracy points over base, re-graded with the harness of \S\ref{app:setup}). The label-free cross-model reward transfers to MATH-500 for $+8.5$, about $81\%$ of the gold-label gain, confirming the effect is not specific to small policies. At 7B the cross and self rewards are comparable on MATH (a stronger policy's own samples are reliable targets), while cross matches or beats self on every benchmark.}
\label{tab:rft7b}
\centering\small
\begin{tabular}{@{}lccc@{}}
\toprule
\textbf{Reward (accuracy $\Delta$ over base)} & \textbf{MATH-500 (transfer)} & \textbf{GSM8K (train dist.)} & \textbf{GPQA (OOD)} \\
\midrule
self-consistency (label-free)      & $+9.0$ & $+3.7$ & $\mathbf{+3.3}$ \\
\rowcolor{rowhl}
cross-model consensus (label-free) & $+8.5$ & $\mathbf{+5.7}$ & $+2.7$ \\
gold (oracle, uses labels)         & $\mathbf{+10.5}$ & $+1.7$ & $+2.7$ \\
\bottomrule
\end{tabular}
\end{table}

\section{Selective-Prediction Curves}
\label{app:selective}
\begin{figure}[t]
\centering
\includegraphics[width=\linewidth]{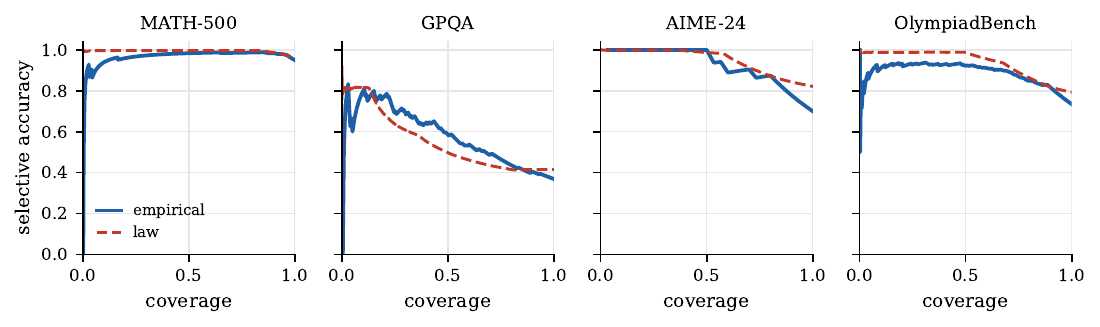}
\caption{\textbf{Selective accuracy vs.\ coverage} on MATH-500, GPQA, AIME-24, and OlympiadBench, empirical (solid) vs.\ the parameter-free law (dashed). Accepting only the most-agreed problems raises accuracy sharply, and the law predicts the curve with no per-benchmark fitting. The empirical curve is noisy at low coverage on GPQA, where few problems reach full unanimity.}
\label{fig:selective}
\end{figure}

Figure~\ref{fig:selective} shows the empirical selective-prediction curve (accuracy versus coverage, accepting the most-agreed problems first) against the law's prediction, per benchmark. The law tracks the curve closely on the saturated and competition-math sets and captures the qualitative shape on GPQA, where the low-coverage regime is noisy because few problems reach full unanimity. These curves are the operating characteristic for using consensus as an abstention or routing signal: they specify the coverage at which a target accuracy is met. Table~\ref{tab:operating} reads two representative operating points (unanimous and three-quarters-majority acceptance) off these curves for a four-model cross-family panel.

\begin{table}[ht]
\caption{\textbf{Agreement as an abstention dial} (four-model cross-family panel answering directly). Answering only when the panel agrees raises accuracy far above the answer-all rate: a unanimous panel is $99.5\%$ correct on MATH-500 (at $85\%$ coverage) and $100\%$ on AIME. On GPQA and MMLU-Pro unanimous accuracy plateaus below one ($0.769$, $0.702$), capped by the shared-error floor.}
\label{tab:operating}
\centering\small
\setlength{\tabcolsep}{5pt}
\begin{tabular}{@{}lccccc@{}}
\toprule
& & \multicolumn{2}{c}{\textbf{Unanimous ($g{=}1$)}} & \multicolumn{2}{c}{\textbf{Majority ($g{\ge}3/4$)}} \\
\cmidrule(lr){3-4}\cmidrule(lr){5-6}
\textbf{Benchmark} & \textbf{Answer-all acc.} & \textbf{coverage} & \textbf{accuracy} & \textbf{coverage} & \textbf{accuracy} \\
\midrule
GSM8K          & $0.973$ & $0.935$ & $0.985$ & $0.988$ & $0.979$ \\
MATH-500       & $0.956$ & $0.850$ & $0.995$ & $0.942$ & $0.985$ \\
OlympiadBench  & $0.734$ & $0.534$ & $0.925$ & $0.718$ & $0.888$ \\
AIME-2024      & $0.700$ & $0.333$ & $1.000$ & $0.500$ & $1.000$ \\
AIME-2025      & $0.433$ & $0.200$ & $1.000$ & $0.400$ & $1.000$ \\
GPQA           & $0.369$ & $0.131$ & $0.769$ & $0.313$ & $0.694$ \\
MMLU-Pro       & $0.458$ & $0.480$ & $0.702$ & $0.777$ & $0.559$ \\
\bottomrule
\end{tabular}
\end{table}

\section{Code Domain: Behavioral Consensus on HumanEval+}
\label{app:code}
For code we cannot match answer strings, so we define agreement behaviorally. For each problem we extract a shared battery of test inputs from the benchmark harness (mean $24.5$ inputs), execute each panel model's program on them, and cluster programs by identical output vectors. The consensus program is a representative of the largest class. Correctness is decided by the held-out unit tests. Per-model pass@1 on the four-model panel ranges $0.811$--$0.933$, and behavioral cross-model consensus reaches $0.909$, near the best single member ($0.933$). The reliability signal is what carries over from the math setting: behavioral agreement predicts correctness at AUROC $0.748$, and the $147/164$ problems on which the panel is behaviorally unanimous pass at $0.952$ versus $0.529$ on the $17$ split problems. This demonstrates the consensus signal in a modality with no notion of a matched answer string.

\section{Agreement-Gated Cascade: Details}
\label{app:cascade}
Because cross-model disagreement flags the hard problems, agreement doubles as a routing rule. We run a cheap two-model panel (the two open-weight members) on every problem. If the two agree we accept their answer at a cost of two model calls, and only on disagreement do we escalate to the full four-model panel (four calls). Table~\ref{tab:cascade} reports the cascade against the always-cheap and always-full panels, and against self-consistency at the cascade's matched average cost. Empirically the cascade matches the full panel's accuracy on every benchmark: on the problems where the cheap pair agrees its answer coincides with the full-panel consensus, and on the rest the cascade defers to that full panel, so escalation loses nothing. Yet its cost remains close to the cheap panel's: $2.2$ calls on MATH-500 and $2.1$ on GSM8K, rising to $3.2$--$3.4$ on AIME/GPQA where more problems are genuinely contested. The escalation rate is itself a difficulty estimate ($3\%$ GSM8K, $11\%$ MATH, $60$--$70\%$ AIME/GPQA). At the same average budget, self-consistency is much weaker (AIME-2024 $0.367$ vs.\ the cascade's $0.700$, GPQA $0.288$ vs.\ $0.369$), because spending the budget on more samples of one model cannot escape that model's correlated errors. A three-model cheap tier gives no further benefit over the two-model tier, so the minimal cascade is the efficient one.

\begin{table}[ht]
\caption{\textbf{Agreement-gated cascade} (accept a unanimous two-model panel, escalating splits to the full four-model panel). The cascade matches the full panel's accuracy at an average cost near the two-model panel, and beats self-consistency at the same average cost (last column). Cost is model calls per problem, and the escalation rate self-scales with difficulty.}
\label{tab:cascade}
\centering\small
\setlength{\tabcolsep}{5pt}
\begin{tabular}{@{}lcccccc@{}}
\toprule
\textbf{Benchmark} & \textbf{Cheap ($M{=}2$)} & \textbf{Full ($M{=}4$)} & \textbf{Cascade} & \textbf{Escalate} & \textbf{Avg.\ cost} & \textbf{SC @cost} \\
\midrule
GSM8K     & $0.961$ & $0.973$ & $0.973$ & $3\%$  & $2.07$ & $0.961$ \\
MATH-500  & $0.920$ & $0.956$ & $0.956$ & $11\%$ & $2.22$ & $0.918$ \\
GPQA      & $0.313$ & $0.369$ & $0.369$ & $70\%$ & $3.40$ & $0.288$ \\
AIME-24   & $0.400$ & $0.700$ & $0.700$ & $60\%$ & $3.20$ & $0.367$ \\
AIME-25   & $0.233$ & $0.433$ & $0.433$ & $63\%$ & $3.27$ & $0.300$ \\
\bottomrule
\end{tabular}
\end{table}

\begin{figure}[ht]
\centering
\includegraphics[width=1.0\linewidth]{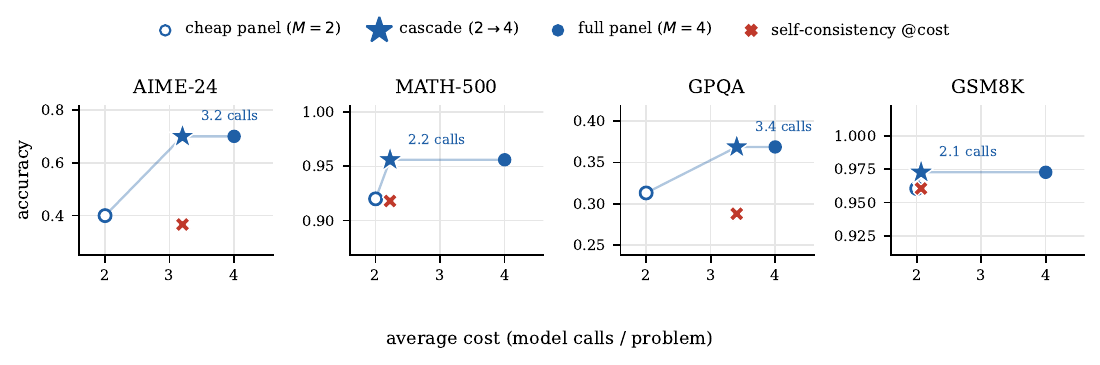}
\caption{\textbf{Agreement-gated cascade.} Accuracy vs.\ average cost (model calls/problem). A cheap two-model panel (hollow) is accepted when unanimous and escalated to the full four-model panel (filled) on disagreement. The cascade (star) attains full-panel accuracy at near-cheap cost. At the cascade's matched cost, self-consistency ($\times$) is far weaker on the unsaturated benchmarks. Escalation self-scales with difficulty, so the curve is steepest where headroom is largest.}
\label{fig:cascade}
\end{figure}

\section{Generator-Robustness of the Verifier}
\label{app:gen2}
To confirm the verifier comparison is not specific to the Qwen3-235B generator of Table~\ref{tab:verifier}, we repeat the full seven-benchmark comparison with a different-family generator, DeepSeek-V3.2, using the remaining cross-family models as the selection panel and DeepSeek-V3.2 itself as the LLM-verifier (Table~\ref{tab:gen2}). The results reproduce those of Table~\ref{tab:verifier}. Cross-model consensus beats the single-model LLM-verifier on \emph{all seven} benchmarks (paired bootstrap $p{\le}0.016$ throughout), reconfirming that a model cannot adjudicate its own candidates regardless of which model generates them. It beats self-consistency significantly on every unsaturated benchmark with headroom (AIME-2024 $+20.0$, $p{=}0.001$; OlympiadBench $+4.8$, $p{<}0.001$; GPQA $+6.6$, $p{=}0.006$; MMLU-Pro $+3.3$, $p{<}0.001$) and ties it on the saturated math sets (MATH-500, GSM8K) and the small-sample AIME-2025, exactly the no-headroom regimes where it also only tied in the main table. The selection-signal advantage is therefore a property of the cross-model signal and does not depend on which generator supplies the pool.

\begin{table}[ht]
\caption{\textbf{Generator-robustness}: the seven-benchmark verifier comparison with a DeepSeek-V3.2 generator (cf.\ Table~\ref{tab:verifier}, Qwen3-235B generator). Cross-model consensus beats the LLM-verifier on all seven ($p{\le}0.016$) and beats self-consistency on every unsaturated benchmark (AIME-24 $p{=}0.001$, OlympiadBench/MMLU-Pro $p{<}0.001$, GPQA $p{=}0.006$), tying only on saturated math and small-sample AIME-2025. The advantage does not depend on the generator.}
\label{tab:gen2}
\centering\footnotesize
\setlength{\tabcolsep}{4pt}
\begin{tabular}{@{}lccccccc@{}}
\toprule
\textbf{Selector} & \textbf{AIME-24} & \textbf{AIME-25} & \textbf{MATH} & \textbf{Olymp.} & \textbf{GPQA} & \textbf{MMLU-Pro} & \textbf{GSM8K} \\
\midrule
Self-consistency      & $63.3$ & $\mathbf{53.3}$ & $94.4$ & $72.1$ & $28.8$ & $42.5$ & $\mathbf{97.2}$ \\
LLM-verifier          & $63.3$ & $\mathbf{53.3}$ & $91.0$ & $70.0$ & $25.3$ & $40.4$ & $95.8$ \\
\rowcolor{rowhl}
Cross-model consensus & $\mathbf{83.3}$ & $\mathbf{53.3}$ & $\mathbf{95.0}$ & $\mathbf{76.9}$ & $\mathbf{35.4}$ & $\mathbf{45.8}$ & $\mathbf{97.2}$ \\
Oracle (upper bound)  & $83.3$ & $63.3$ & $98.6$ & $84.7$ & $52.0$ & $51.8$ & $97.9$ \\
\bottomrule
\end{tabular}
\end{table}

\paragraph{Fully open-weight panel.} The default panel includes one proprietary member (Claude Sonnet~4.6), so we verify the signal survives when \emph{every} model is open-weight. Dropping the proprietary peer leaves a three-model open panel (Qwen3-235B generator plus the DeepSeek-V3.2 and Kimi-K2.5 peers) evaluated on the same candidate pools (Table~\ref{tab:openpanel}). It reproduces the effect: cross-model consensus beats self-consistency on every unsaturated benchmark (AIME-2024 $+16.7$, OlympiadBench $+5.1$, GPQA $+3.6$, MMLU-Pro $+1.4$) and tracks the default four-model panel within about two points everywhere, in fact exceeding it on GPQA ($36.9$ vs.\ $35.9$). The advantage is therefore not an artifact of the one closed model: a panel a practitioner can run entirely from open weights recovers essentially the full gain, so the method and its reproducibility do not depend on any proprietary API.

\begin{table}[ht]
\caption{\textbf{Fully open-weight panel} (Qwen3-235B generator; peers \{DeepSeek-V3.2, Kimi-K2.5\}, the proprietary Claude member removed), on the candidate pools of Table~\ref{tab:verifier}. The open three-model panel beats self-consistency on every unsaturated benchmark and matches the default four-model panel within $\approx2$ points, exceeding it on GPQA. The cross-model advantage does not depend on a proprietary model.}
\label{tab:openpanel}
\centering\footnotesize
\setlength{\tabcolsep}{4pt}
\begin{tabular}{@{}lccccccc@{}}
\toprule
\textbf{Selector} & \textbf{AIME-24} & \textbf{AIME-25} & \textbf{MATH} & \textbf{Olymp.} & \textbf{GPQA} & \textbf{MMLU-Pro} & \textbf{GSM8K} \\
\midrule
Self-consistency          & $36.7$ & $33.3$ & $95.2$ & $65.4$ & $33.3$ & $43.8$ & $96.7$ \\
Cross-model (default, 4)  & $\mathbf{56.7}$ & $33.3$ & $\mathbf{96.6}$ & $\mathbf{72.6}$ & $35.9$ & $\mathbf{46.2}$ & $\mathbf{97.2}$ \\
\rowcolor{rowhl}
Cross-model (open, 3)     & $53.3$ & $\mathbf{40.0}$ & $96.4$ & $70.5$ & $\mathbf{36.9}$ & $45.2$ & $97.1$ \\
\bottomrule
\end{tabular}
\end{table}

\paragraph{Frontier-model panel.} To test whether the signal survives near-state-of-the-art members, we replace the selection panel with three of the strongest available reasoning models (Claude Opus~4.8, Kimi-K2-Thinking, Claude Sonnet~4.6), keeping the Qwen3-235B candidate pool fixed (Table~\ref{tab:frontier}). The jury remains the strongest non-oracle selector on every unsaturated benchmark, comparable to the default panel of Table~\ref{tab:verifier}: it beats self-consistency on OlympiadBench ($+8.3$), MMLU-Pro ($+2.8$), GPQA ($+2.6$), and AIME-2024 ($+20.0$), and the single-model LLM-verifier throughout. Cross-model decorrelation is therefore a consequence of independent training and does not require weak members: even frontier models err differently enough that their agreement carries signal.

\begin{table}[ht]
\caption{\textbf{Frontier-model jury}: selection panel = \{Claude Opus~4.8, Kimi-K2-Thinking, Claude Sonnet~4.6\} on the Qwen3-235B candidate pool. The jury remains the strongest non-oracle selector and is comparable to the default panel (Table~\ref{tab:verifier}), so the cross-model advantage does not depend on using mid-tier members.}
\label{tab:frontier}
\centering\footnotesize
\setlength{\tabcolsep}{4pt}
\begin{tabular}{@{}lccccccc@{}}
\toprule
\textbf{Selector} & \textbf{AIME-24} & \textbf{AIME-25} & \textbf{MATH} & \textbf{Olymp.} & \textbf{GPQA} & \textbf{MMLU-Pro} & \textbf{GSM8K} \\
\midrule
Self-consistency      & $36.7$ & $33.3$ & $95.2$ & $65.4$ & $33.3$ & $43.8$ & $96.7$ \\
LLM-verifier          & $30.0$ & $33.3$ & $93.2$ & $62.2$ & $26.3$ & $43.9$ & $95.9$ \\
\rowcolor{rowhl}
Cross-model (frontier) & $\mathbf{56.7}$ & $\mathbf{43.3}$ & $\mathbf{96.4}$ & $\mathbf{73.7}$ & $\mathbf{35.9}$ & $\mathbf{46.6}$ & $\mathbf{97.6}$ \\
Oracle (upper bound)   & $56.7$ & $43.3$ & $98.6$ & $79.1$ & $53.0$ & $48.9$ & $97.9$ \\
\bottomrule
\end{tabular}
\end{table}

\section{Additional Verifier Results}
\label{app:verifier}
Table~\ref{tab:verifier} reports point accuracies. The oracle-gap-captured statistic (fraction of the achievable selection improvement realized) is $100\%$ for cross-model on AIME-2024 and is near zero or negative for the single-model LLM-verifier on every benchmark, quantifying the claim that a model cannot adjudicate its own candidates. The benchmark where cross-model consensus does not separate from baselines is GSM8K, whose pool is saturated (oracle $97.9$), exactly where a selection signal is expected to be inert.

\paragraph{Scope of the single-model verifier baseline.} Our single-model LLM-verifier is a pointwise scorer (one model rates each candidate $0$--$10$; \S\ref{app:setup}), the standard generative-verifier setup~\citep{zhang2025genrm}. A more elaborate non-trained judge (multi-criteria or system-2 prompting, pairwise tournaments, or debate) could narrow its gap to the jury, and we do not claim to have bracketed the strongest possible prompted judge. Two points bound how much this matters. First, the jury's decisive comparison is against \emph{trained} verifiers (Table~\ref{tab:prm}), which it matches or exceeds, and a prompted judge is a weaker class of method than those. Second, and more fundamentally, any single-model judge, however prompted, scores candidates with the \emph{same} model whose blind spots produced the errors, the within-model limitation the LLM-jury is designed to escape by using independently trained graders. A stronger prompt can sharpen a judge's scores but cannot give one model access to another's decorrelated errors, which is the signal the jury measures. We therefore expect better judge prompting to raise the baseline without closing the structural gap, and leave a systematic judge-prompting sweep to future work.

\paragraph{The verifier advantage is not an artifact of the candidate count.} Table~\ref{tab:verifier} fixes the pool at $N{=}12$. To check the advantage does not hinge on that choice, we subsample each pool to $N\in\{4,8,12\}$ (first $N$ candidates) and recompute the selectors (Table~\ref{tab:nablation}). Cross-model consensus beats self-consistency at \emph{every} $N$ on all five benchmarks, and on AIME-2024 the margin \emph{grows} monotonically with $N$ ($+6.6$ at $N{=}4$, $+13.3$ at $N{=}8$, $+20.0$ at $N{=}12$) because a larger pool is more likely to contain the answer the independent panel points to, which self-consistency's within-model vote keeps missing. The signal is a property of cross-model agreement and is robust to the choice of $N$.

\begin{table}[ht]
\caption{\textbf{Candidate-count ablation} (Qwen3-235B generator; pools subsampled to the first $N$ candidates). Cross-model consensus ($+\Delta$ over self-consistency) wins at every $N$ on every benchmark, and the gain grows with $N$ on the high-headroom AIME-2024. The verifier advantage is not an artifact of $N{=}12$.}
\label{tab:nablation}
\centering\small
\setlength{\tabcolsep}{5pt}
\begin{tabular}{@{}lccc|ccc|ccc@{}}
\toprule
& \multicolumn{3}{c}{\textbf{$N{=}4$}} & \multicolumn{3}{c}{\textbf{$N{=}8$}} & \multicolumn{3}{c}{\textbf{$N{=}12$}} \\
\cmidrule(lr){2-4}\cmidrule(lr){5-7}\cmidrule(lr){8-10}
\textbf{Benchmark} & \textbf{SC} & \textbf{Cross} & \textbf{$\Delta$} & \textbf{SC} & \textbf{Cross} & \textbf{$\Delta$} & \textbf{SC} & \textbf{Cross} & \textbf{$\Delta$} \\
\midrule
AIME-2024     & $36.7$ & $43.3$ & $+6.6$ & $36.7$ & $50.0$ & $+13.3$ & $36.7$ & $56.7$ & $+20.0$ \\
MATH-500      & $94.4$ & $95.6$ & $+1.2$ & $94.8$ & $96.0$ & $+1.2$ & $95.2$ & $96.6$ & $+1.4$ \\
OlympiadBench & $64.4$ & $69.4$ & $+5.0$ & $64.8$ & $71.2$ & $+6.4$ & $65.4$ & $72.6$ & $+7.2$ \\
GPQA          & $28.3$ & $33.3$ & $+5.0$ & $30.8$ & $34.8$ & $+4.0$ & $33.3$ & $35.9$ & $+2.6$ \\
MMLU-Pro      & $44.1$ & $45.8$ & $+1.7$ & $44.3$ & $45.9$ & $+1.6$ & $43.8$ & $46.2$ & $+2.4$ \\
\bottomrule
\end{tabular}
\end{table}

\paragraph{Excluding the generator, and weighting the vote, both leave the signal unchanged.} Two natural variants probe the selection rule (Table~\ref{tab:ablate}). \emph{Generator-excluded:} in the default rule the generator enters only as a tie-break on pool frequency, never as a voting juror (the verification channel is the three non-generator panel members; \S\ref{app:setup}). Removing even that tie-break, so that candidates are scored purely by the independent panel's agreement, changes accuracy by at most $0.7$ points on any benchmark, so generation and verification are already decoupled and the result does not depend on the generator judging its own pool. \emph{Accuracy-weighted voting:} weighting each member's vote by its measured per-model accuracy (in place of one-model-one-vote) does not help: it is within $1.1$ points everywhere and slightly \emph{worse} on GPQA ($34.8$ vs.\ $35.9$), because the signal is carried by \emph{which} answers independently trained models agree on, and is largely insensitive to the strength of any single member. Plain unweighted majority is therefore the right rule, and the method needs no per-model tuning.

\begin{table}[ht]
\caption{\textbf{Selection-rule ablations.} Removing the generator from the tie-break (pure independent-panel agreement) and weighting votes by per-model accuracy both leave selection accuracy within $\approx1$ point of the default unweighted rule (Qwen3-235B pools, cf.\ Table~\ref{tab:verifier}). Generation and verification are already decoupled, and unweighted majority needs no learned weights.}
\label{tab:ablate}
\centering\small
\setlength{\tabcolsep}{6pt}
\begin{tabular}{@{}lccccccc@{}}
\toprule
\textbf{Selection rule} & \textbf{AIME-24} & \textbf{AIME-25} & \textbf{MATH} & \textbf{Olymp.} & \textbf{GPQA} & \textbf{MMLU-Pro} & \textbf{GSM8K} \\
\midrule
\rowcolor{rowhl}
Default (unweighted)    & $56.7$ & $40.0$ & $96.6$ & $72.6$ & $35.9$ & $46.2$ & $97.2$ \\
Generator-excluded      & $56.7$ & $40.0$ & $96.4$ & $73.3$ & $35.9$ & $46.3$ & $97.3$ \\
Accuracy-weighted       & $56.7$ & $40.0$ & $96.4$ & $72.7$ & $34.8$ & $46.3$ & $97.3$ \\
\bottomrule
\end{tabular}
\end{table}

\paragraph{The self-check gate changes one cell and does not create the advantage.} The trained-verifier comparison (Table~\ref{tab:prm}) trusts a verifier's scores only when they separate correct from incorrect candidates at pooled per-candidate AUROC $\ge 0.6$, falling back to self-consistency otherwise. To confirm this gate is not what produces the jury's advantage, we recompute the four verifiers' selection accuracy both gated and ungated (Table~\ref{tab:gate}). Of the sixteen verifier--benchmark cells, fifteen clear the gate (AUROC $0.60$--$0.99$) and are therefore \emph{identical} gated and ungated. The gate binds on exactly one cell, ThinkPRM on GPQA (AUROC $0.556$), where it raises a $26.8$ to the self-consistency $33.3$. Crucially, the jury's ranking is unchanged either way: even \emph{ungated}, the free jury ($0.966$ on MATH-500, $0.359$ on GPQA) is at least as strong as every trained verifier on every benchmark, so the head-to-head conclusion does not depend on the gate or its threshold. The gate is a fairness provision for the trained verifiers (it can only help them, by discarding their unreliable scores) and does not itself produce the jury's lead. The threshold itself is not load-bearing: reading the AUROC column of Table~\ref{tab:gate}, lowering the gate to $0.55$ admits every cell (the minimum AUROC is $0.556$), recovering the ungated numbers, while raising it to $0.65$ additionally gates the three GPQA verifiers below that value (Qwen-PRM-7B/72B and ThinkPRM) up to the self-consistency $33.3$. Under either threshold the jury remains the strongest selector on all four benchmarks (its GPQA $35.9$ still exceeds the gated $33.3$). No threshold in $[0.55,0.65]$ changes the head-to-head conclusion.

\begin{table}[ht]
\caption{\textbf{Self-check gate sensitivity.} Pooled per-candidate AUROC and \emph{ungated} selection accuracy for each trained verifier (cf.\ Table~\ref{tab:prm}). Fifteen of sixteen cells clear the AUROC $\ge 0.6$ gate and are unchanged by it. The gate binds only on ThinkPRM/GPQA ($\dagger$: AUROC $0.556$, gated up to the self-consistency $33.3$). The jury (bottom, agreement-based, no score to gate) is the strongest selector on every benchmark with or without the gate, so the comparison does not hinge on the threshold.}
\label{tab:gate}
\centering\small
\setlength{\tabcolsep}{5pt}
\begin{tabular}{@{}lcccccccc@{}}
\toprule
& \multicolumn{2}{c}{\textbf{MATH-500}} & \multicolumn{2}{c}{\textbf{AIME-24}} & \multicolumn{2}{c}{\textbf{AIME-25}} & \multicolumn{2}{c}{\textbf{GPQA}} \\
\cmidrule(lr){2-3}\cmidrule(lr){4-5}\cmidrule(lr){6-7}\cmidrule(lr){8-9}
\textbf{Verifier} & \textbf{AUROC} & \textbf{acc} & \textbf{AUROC} & \textbf{acc} & \textbf{AUROC} & \textbf{acc} & \textbf{AUROC} & \textbf{acc} \\
\midrule
Qwen2.5-PRM-7B  & $0.83$ & $92.8$ & $0.77$ & $33.3$ & $0.77$ & $33.3$ & $0.61$ & $29.3$ \\
Qwen2.5-PRM-72B & $0.86$ & $94.6$ & $0.90$ & $50.0$ & $0.88$ & $33.3$ & $0.60$ & $32.3$ \\
AceMath-72B-RM  & $0.78$ & $95.6$ & $0.99$ & $56.7$ & $0.98$ & $36.7$ & $0.69$ & $30.3$ \\
ThinkPRM-14B    & $0.74$ & $93.6$ & $0.81$ & $33.3$ & $0.80$ & $36.7$ & $\mathbf{0.56}$ & $26.8^{\dagger}$ \\
\rowcolor{rowhl}
Cross-model     & --- & $\mathbf{96.6}$ & --- & $\mathbf{56.7}$ & --- & $\mathbf{40.0}$ & --- & $\mathbf{35.9}$ \\
\bottomrule
\end{tabular}
\end{table}
\end{document}